\documentclass[10pt,journal,compsoc]{IEEEtran}
\ifCLASSOPTIONcompsoc
  \usepackage[nocompress]{cite}
\else
  \usepackage{cite}
\fi
\ifCLASSINFOpdf
  \usepackage[pdftex]{graphicx}
\else
\fi
\usepackage{amsmath}
\DeclareMathOperator*{\qaplibbur}{\underline{\texttt{bur}}}
\DeclareMathOperator*{\qaplibtwentysix}{\underline{\texttt{26}}}
\DeclareMathOperator*{\qapliba}{\underline{\texttt{a}}}

\usepackage{stfloats}

\usepackage[vlined,boxed,commentsnumbered,ruled,linesnumbered]{algorithm2e}
\usepackage{multirow}
\usepackage{amssymb}
\usepackage{comment}
\usepackage{subfigure}
\usepackage{color, colortbl}
\usepackage{hyperref}
\usepackage{textcomp}

\definecolor{Gray}{gray}{0.9}

\begin{document}
%
% paper title
% Titles are generally capitalized except for words such as a, an, and, as,
% at, but, by, for, in, nor, of, on, or, the, to and up, which are usually
% not capitalized unless they are the first or last word of the title.
% Linebreaks \\ can be used within to get better formatting as desired.
% Do not put math or special symbols in the title.
\title{Neural Graph Matching Network: Learning Lawler\textquotesingle s Quadratic Assignment Problem with Extension to Hypergraph and Multiple-graph Matching}
%
%
% author names and IEEE memberships
% note positions of commas and nonbreaking spaces ( ~ ) LaTeX will not break
% a structure at a ~ so this keeps an author's name from being broken across
% two lines.
% use \thanks{} to gain access to the first footnote area
% a separate \thanks must be used for each paragraph as LaTeX2e's \thanks
% was not built to handle multiple paragraphs
%
%
%\IEEEcompsocitemizethanks is a special \thanks that produces the bulleted
% lists the Computer Society journals use for "first footnote" author
% affiliations. Use \IEEEcompsocthanksitem which works much like \item
% for each affiliation group. When not in compsoc mode,
% \IEEEcompsocitemizethanks becomes like \thanks and
% \IEEEcompsocthanksitem becomes a line break with idention. This
% facilitates dual compilation, although admittedly the differences in the
% desired content of \author between the different types of papers makes a
% one-size-fits-all approach a daunting prospect. For instance, compsoc 
% journal papers have the author affiliations above the "Manuscript
% received ..."  text while in non-compsoc journals this is reversed. Sigh.

\author{Runzhong~Wang,~\IEEEmembership{Student Member,~IEEE,}
        Junchi~Yan,~\IEEEmembership{Senior Member,~IEEE,}
        and~Xiaokang~Yang,~\IEEEmembership{Fellow,~IEEE}% <-this % stops a space
\IEEEcompsocitemizethanks{\IEEEcompsocthanksitem R. Wang, J. Yan, and X. Yang are with Department of Computer Science and Engineering, and MoE Key Lab of Artificial Intelligence, AI Institute, Shanghai Jiao Tong University, Shanghai, 200240, P.R. China. \protect\\
% note need leading \protect in front of \\ to get a newline within \thanks as
% \\ is fragile and will error, could use \hfil\break instead.
E-mail: \{runzhong.wang, yanjunchi, xkyang\}@sjtu.edu.cn\protect\\
Junchi Yan is the correspondence author.\protect %\\
%Project homepage: https://github.com/thinklab-SJTU/PCA-GM.
}
%\IEEEcompsocthanksitem J. Doe and J. Doe are with Anonymous University.}% <-this % stops an unwanted space
\thanks{Manuscript, under review.}}% received March 25, 2020.}}%; revised August 26, 2015.}}

% note the % following the last \IEEEmembership and also \thanks - 
% these prevent an unwanted space from occurring between the last author name
% and the end of the author line. i.e., if you had this:
% 
% \author{....lastname \thanks{...} \thanks{...} }
%                     ^------------^------------^----Do not want these spaces!
%
% a space would be appended to the last name and could cause every name on that
% line to be shifted left slightly. This is one of those "LaTeX things". For
% instance, "\textbf{A} \textbf{B}" will typeset as "A B" not "AB". To get
% "AB" then you have to do: "\textbf{A}\textbf{B}"
% \thanks is no different in this regard, so shield the last } of each \thanks
% that ends a line with a % and do not let a space in before the next \thanks.
% Spaces after \IEEEmembership other than the last one are OK (and needed) as
% you are supposed to have spaces between the names. For what it is worth,
% this is a minor point as most people would not even notice if the said evil
% space somehow managed to creep in.

% The paper headers
\markboth{Journal of \LaTeX\ Class Files,~Vol.~14, No.~8, August~2015}%
{Shell \MakeLowercase{\textit{et al.}}: Bare Demo of IEEEtran.cls for Computer Society Journals}
\IEEEtitleabstractindextext{%
\begin{abstract}
Graph matching involves combinatorial optimization based on edge-to-edge affinity matrix, which can be generally formulated as Lawler\textquotesingle s Quadratic Assignment Problem (QAP). This paper presents a QAP network directly learning with the affinity matrix (equivalently the association graph) whereby the matching problem is translated into a constrained vertex classification task. The association graph is learned by an embedding network for vertex classification, followed by Sinkhorn normalization and a cross-entropy loss for end-to-end learning. We further improve the embedding model on association graph by introducing Sinkhorn based matching-aware constraint, as well as dummy nodes to deal with unequal sizes of graphs. To our best knowledge, this is one of the first network to directly learn with the general Lawler\textquotesingle s QAP. In contrast, recent deep matching methods focus on the learning of node/edge features in two graphs respectively. We also show how to extend our network to hypergraph matching, and matching of multiple graphs. Experimental results on both synthetic graphs and real-world images show its effectiveness. For pure QAP tasks on synthetic data and QAPLIB benchmark, our method can perform competitively and even surpass state-of-the-art graph matching and QAP solvers with notable less time cost. We provide a project homepage at \href{http://thinklab.sjtu.edu.cn/project/NGM/index.html}{http://thinklab.sjtu.edu.cn/project/NGM/index.html}.  
%Source code is public available as a module of \href{https://github.com/Thinklab-SJTU/ThinkCO/tree/main/ThinkMatch}{https://github.com/Thinklab-SJTU/ThinkCO/tree/main/ThinkMatch}
%All these features are welcomed for real-world applications. 
  
  %Our network has its real-world applicability in that it can be trained by graph pairs for each containing different numbers of nodes.
\end{abstract}

% Note that keywords are not normally used for peerreview papers.
}

% make the title area
\maketitle

% To allow for easy dual compilation without having to reenter the
% abstract/keywords data, the \IEEEtitleabstractindextext text will
% not be used in maketitle, but will appear (i.e., to be "transported")
% here as \IEEEdisplaynontitleabstractindextext when the compsoc 
% or transmag modes are not selected <OR> if conference mode is selected 
% - because all conference papers position the abstract like regular
% papers do.
\IEEEdisplaynontitleabstractindextext
% \IEEEdisplaynontitleabstractindextext has no effect when using
% compsoc or transmag under a non-conference mode.

% For peer review papers, you can put extra information on the cover
% page as needed:
% \ifCLASSOPTIONpeerreview
% \begin{center} \bfseries EDICS Category: 3-BBND \end{center}
% \fi
%
% For peerreview papers, this IEEEtran command inserts a page break and
% creates the second title. It will be ignored for other modes.
\IEEEpeerreviewmaketitle

% In this paper, we propose the first combination supervised end-to-end deep graph matching model adopting graph embedding layers, especially cross-graph embedding.
\IEEEraisesectionheading{\section{Introduction and Preliminaries}\label{sec:intro}}
\IEEEPARstart{G}raph matching (GM) is a fundamental problem which is NP-complete in general~\cite{Gare90NPComplete}. It has various applicability and connection with vision and learning, which involves establishing node correspondences between two graphs based on the node-to-node and edge-to-edge affinity~\cite{ChoECCV10,GoldPAMI96}. %The problem can be generalized to the higher-order case whereby hyperedges and their affinity~\cite{LeeCVPR11,NgocCVPR15,YanCVPR15} are defined for matching. 
This differs from the point-based techniques e.g.\ RANSAC~\cite{FischlerCACM81} and iterative closest point (ICP)~\cite{ZhangIJCV94} without considering edge information.
%without explicit modeling beyond node-to-node affinity.

% for $\mathcal{G}_1$ and $\mathcal{G}_2$
We start with two-graph matching, which can be written as quadratic assignment programming (QAP)~\cite{LoiolaEJOR07}, where $\mathbf{X}\in\mathbb{R}^{n_1\times n_2}$ is a (partial) permutation matrix encoding node-to-node correspondence (with constraints in second line of Eq.~(\ref{eq:lawler_qap})), and $\text{vec}(\mathbf{X})$ is its column-vectorized version:
\begin{equation}\label{eq:lawler_qap}
\begin{split}
&J(\mathbf{X}) = \text{vec}(\mathbf{X})^\top\mathbf{K}\text{vec}(\mathbf{X}) \\
s.t. \quad &\mathbf{X}\in \{0,1\}^{n_1\times n_2}, \mathbf{X}\mathbf{1}_{n_2}= \mathbf{1}_{n_1},  \mathbf{X}^\top\mathbf{1}_{n_1}\leq\mathbf{1}_{n_2}
\end{split}
\end{equation}
Here $\mathbf{1}_n$ means column vector of length $n$ whose elements all equal to 1, and $\mathbf{K}\in \mathbb{R}^{n_1n_2\times n_1n_2}$ is the so-called affinity matrix~\cite{LeordeanuICCV05}. Its diagonal and off-diagonal elements store the node-to-node and edge-to-edge affinities. For graph matching, the objective $J(\mathbf{X})$ is maximized, assuming perfect matching corresponds to the highest affinity score. One popular embodiment of $\mathbf{K}$ is fixed Gaussian kernel with Euclid distance over pairs of edge features~\cite{ChoECCV10,YanPAMI16}:
\begin{equation}\label{eq:gaussian_kernel}
    K_{ia,jb}=\exp\left(\frac{||\mathbf{f}_{ij}-\mathbf{f}_{ab}||^2}{\sigma^2}\right)
\end{equation}
where $\mathbf{f}_{ij}$ is the feature vector of the edge $E_{ij}$ in graph 1, $\mathbf{f}_{ab}$ from edge $E_{ab}$ in graph 2. $K_{ia,jb}$ is indexed by $(an_1+i,bn_1+j)$ under its matrix form.
%which can also incorporate the node similarity when node index $i=a$.  

\begin{table}[tb!]
    \centering
    \caption{Summary of existing learning-free and learning-based methods on two popular formulations of QAP. Note that Lawler\textquotesingle s QAP (Eq.~\ref{eq:lawler_qap}) is most general which incorporates the Koopmans-Beckmann\textquotesingle s QAP (Eq.~\ref{eq:kb_qap}).}
    %\vspace{-10pt}
    \resizebox{0.5\textwidth}{!}
    {
    \begin{tabular}{c|c|c}
    \hline
         & Koopmans-Beckmann\textquotesingle s QAP & Lawler\textquotesingle s QAP \\
    \hline
        learning-free & \cite{edwards1980branch,punnen2013linear,ErdouganCOR07,dokeroglu2016novel} & \cite{HahnEJOR98,LeordeanuICCV05,ChoECCV10,WangPAMI17,KushinskySIAM19}\\
    \hline
        learning-based & \cite{nowak2018revised,WangICCV19} & ours \\
    \hline
    \end{tabular}
    }

    \label{tab:kb_lawler_summary}
\end{table}

Note Eq.~(\ref{eq:lawler_qap}) in literature is called Lawler\textquotesingle s QAP~\cite{LawlerMS63}, which can incorporate other special forms. For instance, the popular Koopmans-Beckmann\textquotesingle s QAP~\cite{KBQAP57} is written by:
\begin{equation}
J(\mathbf{X})=\text{tr}(\mathbf{X}^\top\mathbf{F}_1\mathbf{X}\mathbf{F}_2)+\text{tr}(\mathbf{K}_p^\top\mathbf{X})
\label{eq:kb_qap}
\end{equation}
%of $\mathcal{G}_1$, $\mathcal{G}_2$
where $\mathbf{F}_1\in \mathbb{R}^{n_1\times n_1}$, $\mathbf{F}_2\in \mathbb{R}^{n_2\times n_2}$ are weighted adjacency matrices. $\mathbf{K}_p\in\mathbb{R}^{n_1\times n_2}$ is node-to-node affinity matrix. Its connection to Lawler\textquotesingle s QAP becomes clear by letting $\mathbf{K}=\mathbf{F}_2\otimes_{\mathcal{K}}\mathbf{F}_1$ ($\otimes_{\mathcal{K}}$ means Kronecker product).

\begin{comment}
Another popular formulation is the so-called factorized graph matching model \cite{ZhouPAMI16}, which shows how to factorize affinity matrix $\mathbf{K}$ as a Kronecker product of smaller matrices. For concise, here we write the undirected version~\cite{ZhouPAMI16}:
\begin{align}
&\mathbf{K}=(\mathbf{H}_2\otimes\mathbf{H}_1)\text{diag}(\text{vec}(\mathbf{L}))(\mathbf{H}_j\otimes\mathbf{H}_i)^\top\\\notag
\text{where}\quad&\mathbf{H}_k=[\mathbf{G}_k,\mathbf{I}_{n_k}]\in\{0,1\}^{n_k\times(m_k+n_k)},\quad k=i,j\\\notag
&\mathbf{L}= \left[\begin{array}{ll} \mathbf{K}^{q} & -\mathbf{K}^{q}\mathbf{G}_{j}^\top\\
-\mathbf{G}_{i}\mathbf{K}^{q}  & \mathbf{G}_{i}\mathbf{K}^{q}\mathbf{G}_{j}^\top+\mathbf{K}^{p}\end{array}\right]
\end{align}
where $n_i$ and $m_i$ is the number of nodes and edges in graph $\mathcal{G}_i$ respectively and $\otimes$ is the Kronecker product operation between matrices. $\mathbf{K}^{p}\in\mathbb{R}^{n_i\times n_j}$ denotes the node affinity matrix, and $\mathbf{K}^q\in\mathbb{R}^{m_i\times m_j}$ for the edge affinity matrix. The graph structure is specified by the node-edge incidence matrix $\mathbf{G}\in\mathbb{R}^{n\times m}$ such that the non-zero elements in each column of $\mathbf{G}$ indicate the starting and ending nodes in the corresponding edge. The factorization provides a taxonomy for GM and reveals the connection among several methods. Readers are referred to \cite{ZhouPAMI16} for greater details.
\end{comment}

\begin{figure*}[tb!]
  \centering
  \includegraphics[width=0.96\textwidth]{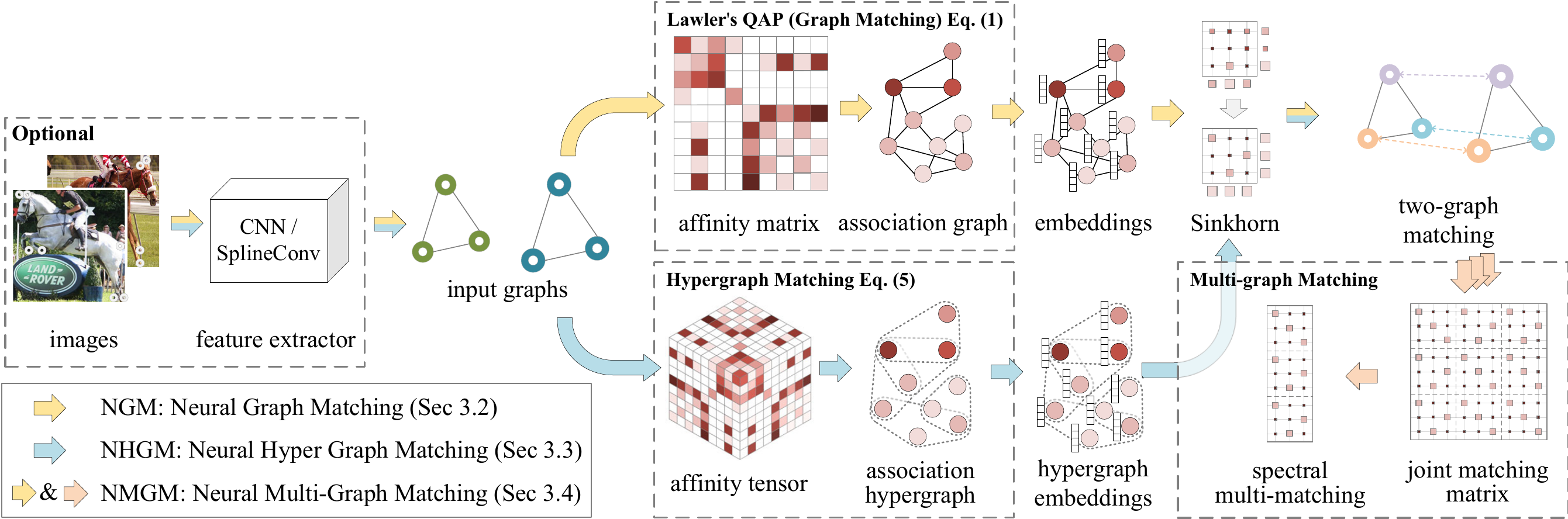}
  %\vspace{-5pt}
 %  \vspace{-10pt}
  \caption{Overview of the proposed neural graph matching pipeline. The method directly handles Lawler\textquotesingle s QAP based on the embedding on the association graph. We further extend it to hypergraph matching by replacing the association graph with an association hypergraph (see Sec.~\ref{sec:nhgm}), as well as to multiple graph matching by differentiable spectral multi-matching (see Sec.~\ref{sec:nmgm}).}
  %\vspace{-10pt}
  \label{fig:overview}
\end{figure*}

As shown in Tab.~\ref{tab:kb_lawler_summary}, traditional learning-free solvers have been extensively studied for both QAP formulations in Eq.~(\ref{eq:lawler_qap}, \ref{eq:kb_qap}), and there exist some recent advances in learning Koopmans-Beckmann\textquotesingle s QAP~\cite{nowak2018revised,WangICCV19}. In this paper, we propose the first learning-based algorithm tackling the most general QAP form -- Lawler\textquotesingle s QAP, and show its generalization to higher-order and multi-graph scenarios.

The above QAP models involve the second-order affinity, and can also be generalized to the higher-order case. A line of works \cite{ChertokPAMI10,DuchennePAMI11,YanCVPR15,ZassCVPR08} adopt tensor marginalization based model for $m$-order ($m\ge 3$) hypergraph matching, resulting in a higher-order assignment problem:
\begin{equation}\label{eq:formulationB}
\begin{split}
&J(\mathbf{x})=\mathbf{H}\otimes_1 \mathbf{x}\otimes_2 \mathbf{x}\ldots\otimes_m \mathbf{x}\\
s.t.\quad &\mathbf{X}\mathbf{1}_{n_2}= \mathbf{1}_{n_1}, \mathbf{X}^\top\mathbf{1}_{n_1}\leq\mathbf{1}_{n_2}
\end{split}
\end{equation}
where $\mathbf{x} = \text{vec}(\mathbf{X}) \in \{0,1\}^{n_1n_2\times 1}$ is the column-vectorized form, and $\mathbf{H}$ is the $m$-order affinity tensor whose $(n_1n_2)^m$ elements record the affinity between two hyperedges, operated by tensor product $\otimes_k$ \cite{LeeCVPR11}:
\begin{equation}
\label{eq:tensor_prod}
    (\mathbf{H} \otimes_k \mathbf{x})_{...,i_{k-1},i_{k+1},...} = \sum_{i_k=1}^{n_1n_2}  \mathbf{H}_{...,i_{k-1},i_k,i_{k+1},...} \cdot \mathbf{x}_{i_k}
\end{equation}
where $\otimes_k$ can be regarded as tensor marginalization at dimension $k$. Details of tensor multiplication can be referred to Sec. 3.1 in \cite{DuchennePAMI11}. Most existing hypergraph matching works assume the affinity tensor is invariant w.r.t. the index of the hyperedge pairs for computational tractability.

%In addition with the above affinity formulations for two-graph matching, there are also a considerable amount of recent efforts on multiple graph matching approaches, where these approaches mostly either transform the problem into a two-graph matching QAP in each iteration \cite{YanPAMI16,yan2015consistency}, or first solve two-graph matching problems to obtain the putative matchings for further post-processing \cite{chen2014near,HuangSGP13,PachauriNIPS13,wang2018multi}.

As discussed above, either graph matching or hypergraph matching problem involves solving a combinatorial optimization problem. However, the objective functions may be biased and even the mathematically optimal solution can depart from the perfect matching in reality, either due to the noise observation or limited modeling capacity, or both. In fact, traditional methods are mostly based on predefined shallow affinity with limited capacity e.g.\ Gaussian kernel with Euclid distance (see Eq.~(\ref{eq:gaussian_kernel})), which has difficulty in providing enough flexibility for real-world data. This issue is partially addressed by affinity-learning based graph matching~\cite{WangICCV19,ZanfirCVPR18,ChoICCV13}. Along this promising direction, in this paper, a novel network based solver is proposed to directly learn Lawler\textquotesingle s QAP whereby the affinity learning is also incorporated, as shown in Fig.~\ref{fig:overview}. This approach is further extended to the case of joint matching of multiple graphs, which has been an important scenario in practice and has received wide attention in literature~\cite{YanPAMI16,yan2015consistency,chen2014near,PachauriNIPS13,WangCVPR18} however learning has not been considered. Also hypergraph matching is enabled in our framework.

%As a common technique to mitigate biased pairwise matching objective in existing learning-free works, multi-graph information is also considered in our end-to-end framework.  shows our overall pipeline.
%However, the learned affinity matrix cannot fully eliminate the bias and can still mislead the matching. 

%Traditional methods mostly assume the affinity is predefined and focus on finding the optimal solution. The limited capacity of a shallow affinity model e.g. Gaussian kernel with Euclid distance (see Eq.~\ref{eq:gaussian_kernel}), cannot provide enough flexibility for real-world data. In other words, the predefined affinity function can be biased and even the mathematically optimal solution can depart from the perfect matching in reality. Such challenges have also been (partially) addressed in multiple graph matching, which tries to enforce cycle-consistency to mitigate local bias~\cite{YanPAMI16,yan2015consistency,PachauriNIPS13}. In this paper, we resort to deep networks to train the affinity function adapted to data. Moreover, our network further learns to solve QAP (and matching of hypergraph as well as multiple graphs) end-to-end in a data-driven manner. % by discovering the underlying pattern of specific problems 
Specifically, the proposed matching nets consist of several learnable layers as detailed in Fig.~\ref{fig:network_structure}: 1) CNN layers taking raw images for node (and edge) feature extraction; 2) Spline convolution layers encoding geometric features to node (and edge) features; 3) affinity metric learning layer for generating the affinity matrix i.e. the association graph; 4) vertex embedding layers using the association graph as input for vertex classification; 5) Sinkhorn net to convert the vertex score matrix into doubly-stochastic matrix. Sinkhorn technique is also adopted in the embedding module to introduce matching constraints; 6) cross-entropy loss layer whose input is the output of the Sinkhorn layer.

Note that the first two components are optional and can be treated as a plugin in the pipeline, and have nothing to do with Lawler\textquotesingle s QAP. In contrast, the peer network \cite{ZanfirCVPR18} only allows for learning of node CNN features on images and their similarity metric i.e. the component 1) and 2). In fact, \cite{ZanfirCVPR18} is inapplicable to learning the QAP model. Our embedding also differs from \cite{ZhangICCV19,WangICCV19,WangCVPR20} as raw individual graphs must be required for embedding in these works. In fact, \cite{nowak2018revised} shows that embedding on individual graphs can deal with some special cases of Koopmans-Beckmann\textquotesingle s QAP, which is also a special case of Lawler\textquotesingle s QAP as discussed above. Direct learning on Lawler\textquotesingle s QAP enables a learning-based solver for real-world combinatorial problems beyond vision, e.g.\ QAPLIB problem instances~\cite{Burkard1997QAPLIB}, which can not be readily handled by previous graph matching learning algorithms.

Furthermore, we devise two generalizations to the above matching network. 1) hypergraph matching by embedding a higher-order affinity tensor; 2) multiple graph matching by devising an end-to-end compatible matching synchronization module by using the popular spectral fusion technique~\cite{PachauriNIPS13,MasetICCV17}. The performance can also be boosted by adopting edge-embedding layers.

\begin{table*}[tb!]
    \centering    
    \caption{Summary of existing literature in learning graph matching based on different types of assignment problems solved by learning, learning modules including CNN, GNN and affinity metric, where GNN embedding is performed and loss functions. KB-QAP abbreviates Koopmans-Beckmann\textquotesingle s QAP in Eq.~(\ref{eq:kb_qap}). Lawler\textquotesingle s QAP in Eq.~(\ref{eq:lawler_qap}) is the most general QAP form and higher-order assignment in Eq.~(\ref{eq:formulationB}) is its higher-order extension.}
    %\vspace{-10pt}
    \resizebox{\textwidth}{!}
    {
\begin{tabular}{r|l|l|l|l|l|l|l}
\hline
method & learned neural net solver & multiple-graph&CNN&GNN module&embedded graph&affinity metric&loss function\\\hline
{Nowak \textit{et al.} \cite{nowak2018revised}} & special case of KB-QAP & {none} & {none}  & {GCN} & {individual graphs} & {inner-product} & {multi-class cross-entropy} \\
\hline
GMN~\cite{ZanfirCVPR18} & none & {none} & VGG16 & none & none & weighted exponential  & pixel offset regression \\\hline
\multirow{2}{*}{Zhang \textit{et al.} \cite{ZhangICCV19}} & \multirow{2}{*}{none} & \multirow{2}{*}{none} & \multirow{2}{*}{none} & message-passing &  \multirow{2}{*}{individual graphs} & \multirow{2}{*}{inner-product} & \multirow{2}{*}{multi-class cross-entropy}\\
& & & &based CMPNN & &\\\hline
PCA-GM~\cite{WangICCV19} \& & \multirow{2}{*}{special case of KB-QAP} & \multirow{2}{*}{none} & \multirow{2}{*}{VGG16} & GCN+cross-graph &  \multirow{2}{*}{individual graphs} & \multirow{2}{*}{weighted exponential} & \multirow{2}{*}{binary cross-entropy}  \\
IPCA-GM~\cite{WangPAMI20} &  & &&  convolution &&\\\hline
\multirow{2}{*}{CIE-H\cite{YuICLR20}} & \multirow{2}{*}{special case of KB-QAP} & \multirow{2}{*}{none} & \multirow{2}{*}{VGG16} & edge embedding+  & \multirow{2}{*}{individual graphs} &  \multirow{2}{*}{weighted exponential} & binary cross-entropy \\
 &&&& cross-graph conv. &&& with Hungarian attention \\\hline
\multirow{2}{*}{LCS~\cite{WangCVPR20}} & special case of Lawler\textquotesingle s & \multirow{2}{*}{none} & \multirow{2}{*}{VGG16} & \multirow{2}{*}{GCN} &  \multirow{2}{*}{association graph} & fully-connected & \multirow{2}{*}{binary cross-entropy}  \\
&  QAP  &&&&& neural network \\\hline
\multirow{2}{*}{BBGM~\cite{RolinekECCV20}} & \multirow{2}{*}{none} & unlearned fixed & \multirow{2}{*}{VGG16} & \multirow{2}{*}{SplineConv} &  \multirow{2}{*}{individual graphs} & \multirow{2}{*}{weighted inner-product} & \multirow{2}{*}{Hamming distance}  \\
& & solver~\cite{SwobodaCVPR19} & &&&  \\\hline
\hline
\multirow{2}{*}{NGM (ours)} & \textbf{general case of Lawler\textquotesingle s} & \multirow{2}{*}{none} & \multirow{2}{*}{VGG16} & matching-aware &  \multirow{2}{*}{association graph} & \multirow{2}{*}{weighted exponential} & \multirow{2}{*}{binary cross-entropy} \\
& \textbf{QAP (most general)} & && GCN & & \\
\hline
%\multirow{2}{*}{NGM+ (ours)} & \textbf{Lawler\textquotesingle s QAP} & \multirow{2}{*}{none} & \multirow{2}{*}{VGG16} & matching-aware & \multirow{2}{*}{association graph} & \multirow{2}{*}{weighted exponential} & \multirow{2}{*}{binary cross-entropy} \\
%& \textbf{(most general)} & &&edge conv. && \\\hline
\multirow{2}{*}{NHGM (ours)} & \multirow{2}{*}{\textbf{higher-order assignment}} & \multirow{2}{*}{none} & \multirow{2}{*}{VGG16} & matching-aware & association hyper- & \multirow{2}{*}{weighted exponential} & \multirow{2}{*}{binary cross-entropy} \\
& & &&hyper-GCN & graph &\\
\hline
\multirow{2}{*}{NMGM (ours)} & \textbf{general case of Lawler\textquotesingle s} & \textbf{end-to-end} & \multirow{2}{*}{VGG16} & matching-aware & \multirow{2}{*}{association graph} &  \multirow{2}{*}{weighted exponential} & \multirow{2}{*}{binary cross-entropy} \\
& \textbf{QAP (most general)} & \textbf{spectral method} & &GCN & &\\
\hline
%\cite{WangArxiv19}& VGG16 & GCN   & binary cross-entropy & weighted exponential  \\\hline
%\cite{JiangArxiv19} & VGG16 & \makecell[l]{Laplacian smoothing/sharpening \\+ cross-graph convolution} & weighted exponential & \makecell[l]{binary cross-entropy\\ + sparsity norm}  \\\hline
%\cite{YuICLR20} & VGG16 & \makecell[l]{Channel independent embedding \\+ cross-graph convolution}  &weighted exponential & \makecell[l]{binary cross-entropy \\with Hungarian attention}\\\hline
%\cite{FeyICLR20}& VGG16 & SplineConv/GIN/cross-GNN &inner-product   & \makecell[l]{BCE+ neighbor consensus} \\
\hline
\multirow{2}{*}{NGM-v2 (ours)} & \textbf{general case of Lawler\textquotesingle s} & \multirow{2}{*}{none} & \multirow{2}{*}{VGG16} & SplineConv+ &  individual graphs & \multirow{2}{*}{weighted inner-product} & \multirow{2}{*}{binary cross-entropy} \\
& \textbf{QAP (most general)} & && match-aware GCN & +asso.\ graph & \\
\hline
\multirow{2}{*}{NHGM-v2 (ours)} & \multirow{2}{*}{\textbf{higher-order assignment}} & \multirow{2}{*}{none} & \multirow{2}{*}{VGG16} & SplineConv+match- &  individual graphs  & \multirow{2}{*}{weighted inner-product} & \multirow{2}{*}{binary cross-entropy} \\
& & &&aware hyper-GCN & +asso.\ hyper-graph &\\
\hline
\multirow{2}{*}{NMGM-v2 (ours)} & \textbf{general case of Lawler\textquotesingle s} & \textbf{end-to-end} & \multirow{2}{*}{VGG16} & SplineConv+  & individual graphs &  \multirow{2}{*}{weighted inner-product} & \multirow{2}{*}{binary cross-entropy} \\
& \textbf{QAP (most general)} & \textbf{spectral method} & & match-aware GCN &  +asso.\ graph &\\
\hline
\end{tabular}
}
%\vspace{-10pt}

    \label{tab:compare_gm_learn}
\end{table*}

The highlights of this paper are summarized as follows:

i) We show how to develop a deep network to directly tackle the (most) general graph matching formulation i.e. Lawler’s Quadratic Assignment Problem beyond vision problems, in the sense of allowing the affinity matrix as the raw input. This is fulfilled by regarding the affinity matrix as an association graph, whose vertices can be embedded by a deep graph neural network (GNN)~\cite{scarselli2008graph} for classification, with a novel matching-aware graph convolution scheme. In contrast, existing works \cite{ChoICCV13,ZanfirCVPR18,ZhangICCV19,WangICCV19} start with individual graphs' node and edge features for affinity learning instead of pairwise affinity encoded in affinity matrix. %Most importantly, we show our learning-based solver is capable to handle biased objective functions, which have not been resolved by previous affinity-learning networks~\cite{ZanfirCVPR18,ZhangICCV19,WangICCV19}.
%We also adopt dummy nodes to handle outliers, which has seldom be addressed in previous deep graph matching methods~\cite{ZanfirCVPR18}.

ii) Our network solver for Lawler’s QAP can be trained either in a supervised setting given ground truth node correspondence from labeled training set (e.g.\ for image matching), or by the final matching score without supervision (e.g.\ for QAPLIB problems).

iii) We extend our second-order graph matching networks to the hypergraph (third-order) matching case. This is fulfilled by building the hyperedge based association hypergraph to replace the second-order one. To our best knowledge, this is the first work for deep learning of hypergraph matching (with explicit treatment on the hyperedges).

iv) We also extend our matching network to the multiple-graph matching case by end-to-end spectral multi-graph matching, with explicit treatment for stabilized learning. To our best knowledge, there is no end-to-end multiple-graph matching neural network in the existing literature.

%Our method notably outperforms the deep network \cite{ZanfirCVPR18}. 
v) Experimental results on synthetic and real-world data show the effectiveness of our devised components. The extended versions for hypergraph matching and multiple-graph matching also show competitive performance. Our model can learn with the Lawler\textquotesingle s QAP as input while state-of-the-art graph matching networks \cite{ZanfirCVPR18,WangICCV19,ZhangICCV19} cannot. This allows for the evaluation of our network on the QAPLIB benchmark directly, which to our best knowledge, is the first test for network-based methods for QAPLIB.

The paper goes as follows. Section~\ref{sec:related} discusses the related work to graph matching and its recent learning based methods. Section~\ref{sec:method} describes the main approach and in Section~\ref{sec:experiment} we discuss the experiments. Section~\ref{sec:conclusion} concludes this paper. Source code is made public available on the project page.

\section{Related Work}
\label{sec:related}
We mainly discuss related works and techniques on learning graph matching~\cite{YanIJCAI20}. Readers are referred to the survey \cite{YanICMR16} for an enlarged overview of the topic of graph matching. Also, a more broad perspective on learning for combinatorial optimization is referred to~\cite{bengio2020machine}.  

%especially for how the affinity matrix or function is modeled and learned. We will also briefly discuss techniques on graph node embedding which will be used in our pipeline for node scoring of the association graph. Related works on permutation learning will also be covered which will be a building block for combinatorial graph matching learning in our approach. 

\subsection{Learning-free Graph Matching Methods}
\textbf{Two-graph matching and QAP.} Lawler\textquotesingle s Quadratic Assignment Problem~\cite{LawlerMS63} is known for its application of matching two graphs by maximizing a quadratic objective function. Traveling salesman problem (TSP) and Koopmans-Beckmann\textquotesingle s QAP are two popular variants from Lawler\textquotesingle s QAP, with their wide range of application beyond vision, e.g.\ economic activities modeled by Koopmans-Beckmann\textquotesingle s QAP~\cite{KBQAP57,Burkard1997QAPLIB}. The most general Lawler\textquotesingle s QAP also refers to two-graph matching in pattern recognition, and is traditionally addressed in a learning-free setting. Classically, small or medium sized problems are tractable by branch-and-bound with dual bound solved approximately~\cite{edwards1980branch,HahnEJOR98}. Modern approximate solvers~\cite{LeordeanuICCV05,ChoECCV10,HungarianMethod09,WangPAMI17,KushinskySIAM19} achieve better accuracy-speed trade-off and thus are more applicable to larger-sized problems. In two-graph matching, most methods focus on seeking approximate solution given fixed affinity model which is often set in simple parametric forms. Euclid distance in node/edge feature space together with a Gaussian kernel to derive a non-negative similarity, is widely used in the above works.

\textbf{Hypergraph matching methods.} 
Going beyond the traditional second-order graph matching, hypergraphs have been built for matching~\cite{ZassCVPR08} and their affinity is usually represented by a tensor to encode the third-order~\cite{YanCVPR15,LeeCVPR11,YanTCYB18} or even higher-order information~\cite{NgocCVPR15}. The advantage is that the model can be more robust against noise at the cost of exponentially increased complexity for both time and space. 

\textbf{Multiple-graph matching methods.} 
It has been recently actively studied for its practical utility against local noise and ambiguity. The hope is that the joint matching of multiple graphs can provide a better venue to fuse the information across graphs, leading to better robustness against local noise and ambiguity. Among the literature, a thread of works~\cite{PachauriNIPS13,chen2014near} first generate the pairwise matching between two graphs via certain two-graph matching solvers, and then impose cycle-consistency on the pairwise matchings to improve the matching accuracy. The other line of methods impose cycle-consistency during the iterative finding of pairwise matchings and usually can achieve better results~\cite{yan2015consistency,YanPAMI16,YanICCV13,YanECCV14}. The online setting for solving multiple graph matching is studied in~\cite{YuECCV18}.

Note both hypergraph or multiple graph matching paradigms try to improve the affinity model either by lifting the affinity order or imposing additional consistency regularization. As shown in the following, another possibly more effective and efficient way is adopting learning to find more adaptive affinity model parameters, or further improving the solver with deep neural networks. % the model capacity by adopting deep neural networks. 

\subsection{Learning-based Graph Matching Methods}
\textbf{Non-deep learning methods.}
The structural SVM based supervised learning method \cite{ChoICCV13} incorporates earlier graph matching learning methods~\cite{TorresaniECCV08,CaetanoPAMI09,LeordeanuIJCV12}. Learning can also be fulfilled by unsupervised~\cite{LeordeanuIJCV12} and semi-supervised~\cite{LeordeanuICCV11}. In these earlier works, no neural network is adopted until the recent seminal work \cite{ZanfirCVPR18}.

%To make the affinity representation model more flexible, a unified while shallow parametric graph structure learning model is devised in a vector form $\Phi(\mathcal{G}_i,\mathcal{G}_j,\pi)$ \cite{ChoICCV13}: $\Phi=[\cdots,s_v(a_u,a_{\pi(u)}),\cdots,s_e(a_{uv},a_{\pi(u)\pi(v)})\cdots]^\top$. By introducing weights on all elements of this feature map, one obtains a score function: $S(\mathcal{G}_i,\mathcal{G}_j,\pi,\beta)=\beta\Phi(\mathcal{G}_i,\mathcal{G}_j,\pi)$ where $\bm{\beta}$ is a weight vector of $s_v$ and $s_u$. In \cite{ChoICCV13} the authors show that the above model can incorporate many previous learning models~\cite{CaetanoPAMI09,LeordeanuIJCV12,TorresaniECCV08}. For instance, \cite{CaetanoPAMI09} adopts a 60-dimensional similarity function $s_v$ for feature points and a binary similarity $s_e$ for edges. \cite{LeordeanuIJCV12} uses a multi-dimensional $s_e$ to measure similarity without taking $s_v$ into consideration. While \cite{TorresaniECCV08} adopts 2-dimensional $s_v$ and $s_e$ functions to model similarity, occlusion, and geometric compatibility. In fact, the learning can be either unsupervised~\cite{LeordeanuIJCV12}, semi-supervised~\cite{LeordeanuICCV11}, or supervised~\cite{CaetanoPAMI09,ChoICCV13,ZanfirCVPR18}.
%As a representative shallow learning model, in experiment the structural SVM based learning method \cite{ChoICCV13} will be compared with our deep learning based approach.

\begin{figure*}[tb!]
  \centering
  \includegraphics[width=0.95\textwidth]{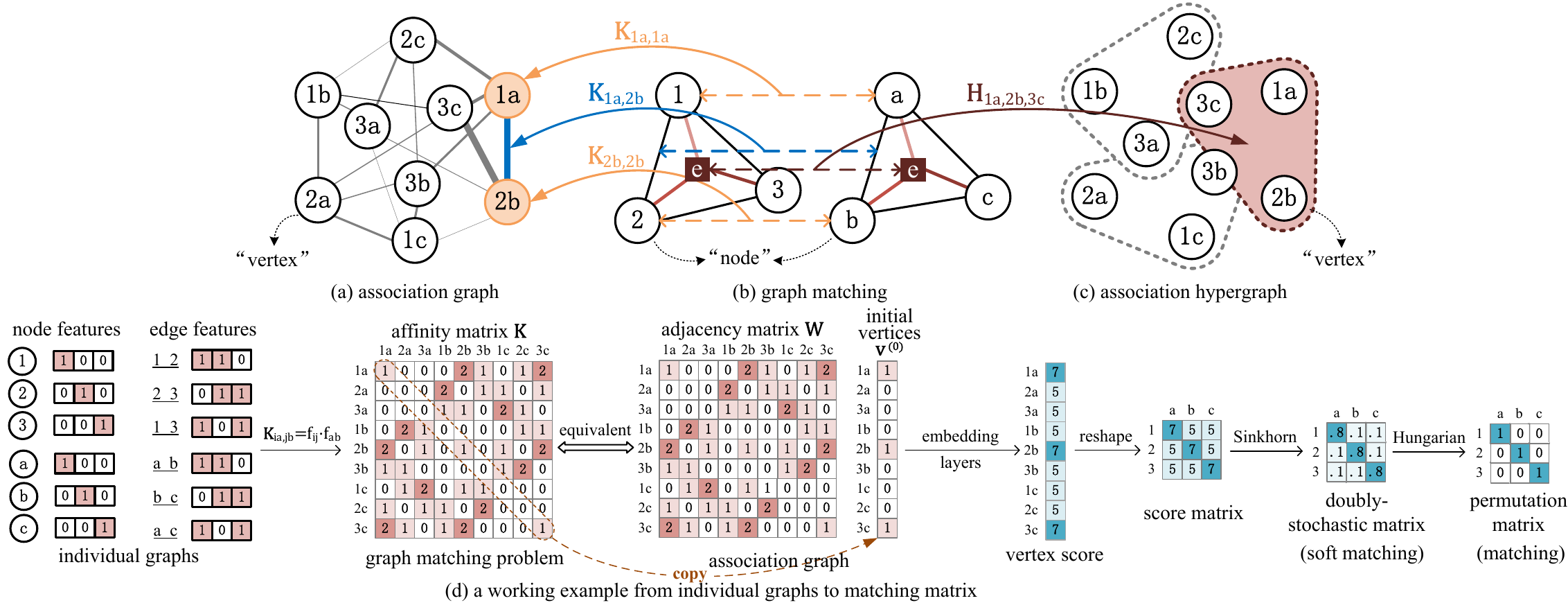}
  %\vspace{-5pt}
  %\in\mathbb{R}^{n_1n_2\times n_1n_2}
  %\in \mathbb{R}^{n_1n_2 \times n_1n_2 \times n_1n_2}
   %\vspace{-20pt}
  \caption{Graphs with affinity matrix $\mathbf{K}$ and affinity tensor $\mathbf{H}$ in (b), w.r.t (a) association graph and (c) association hypergraph. We distinguish the nodes on association (hyper)graph and individual graphs for matching by terms \textit{vertex} and \textit{node} through this paper. The node-to-node matching problem in (b) can therefore be formulated as the vertex classification task on the association graph whose edge weights can be induced by the affinity matrix. Such a perspective is also widely taken in literature for graph matching e.g.\ \cite{ChoECCV10} and hypergraph matching~\cite{LeeCVPR11}. (d) shows a toy working example from individual graphs to final matching result: affinity matrix $\mathbf{K}$ is built from individual graphs, and the matching problem is equivalent to vertex classification on the association graph. Matching-aware embedding and vertex classification are applied on association graph to generate vertex scores, followed with reshaping and Sinkhorn normalization to obtain a double-stochastic matrix i.e. a convex hull of permutation matrix.}
    %\vspace{-10pt}
  \label{fig:association_graph}
\end{figure*}
\textbf{Deep-learning methods.}
A pioneer work \cite{nowak2018revised} considers the alignment of graphs by embedding on individual graphs, which can be regarded a special case of Koopmans-Beckmann\textquotesingle s QAP. Deep learning is recently applied for graph matching on images \cite{ZanfirCVPR18}, whereby convolutional neural network (CNN) is used to extract node features from images followed with spectral matching and CNN is learned using a regression-like node correspondence supervision. This work is improved by introducing GNN to encode structural~\cite{WangICCV19,WangPAMI20} or geometric~\cite{ZhangICCV19} information, with a combinatorial loss based on cross-entropy loss, and Sinkhorn network~\cite{AdamsArxiv11} as adopted in \cite{WangICCV19}. Yu \textit{et al.}~\cite{YuICLR20} extends \cite{WangICCV19} by edge embedding and Hungarian-based attention mechanism to stabilize end-to-end training. %A recent paper BBGM~\cite{RolinekECCV20} propose

We also note one recent important work on learning for graph matching, namely Learning Combinatorial Solver (LCS)~\cite{WangCVPR20}. It follows the line of research~\cite{LeordeanuICCV05,ChoECCV10,TianECCV12} in building an association graph from input images and solves graph matching by vertex classification on association graph. The discussion in \cite{WangCVPR20} is basically restricted to the image matching setting. Though there is much space to generalize their work to our case, while no explicit scheme is given in terms of addressing the most general Lawler\textquotesingle s QAP form. Meanwhile, it is unclear in their method for how to incorporate the matching constraint in the node scoring procedure, which has been addressed by our devised Skinhorn embedding. In fact, our method is directly motivated by developing a general Lawler\textquotesingle s QAP solver, beyond image matching. It also enjoys the flexibility of readily adopting better feature extractors e.g.~\cite{RolinekECCV20} or those beyond  vision. Moreover, our work further develops hypergraph matching and multiple graph matching under the neural learning framework, which are new in literature. All these features are not well explored by LCS~\cite{WangCVPR20}.

As summarized in Tab.~\ref{tab:compare_gm_learn} concerning deep neural network-based graph matching algorithms, one shortcoming of existing graph matching networks is that they cannot directly deal with the most general Lawler\textquotesingle s QAP form which limits their applicability to tasks when no individual graph information is available (see QAPLIB -- \href{http://anjos.mgi.polymtl.ca/qaplib/}{http://anjos.mgi.polymtl.ca/qaplib/}). In contrast, our method can directly work with the affinity matrix, and we further extend to dealing with affinity tensor for hypergraph matching, as well as the setting under multiple graphs. 

\section{Proposed Approaches}\label{sec:method}
In Sec.~\ref{sec:preliminary}, we introduce the preliminary concept association graph, on which our methods are based. In Sec.~\ref{sec:ngm}, we present Neural Graph Matching~(NGM) network, which can solve Lawler\textquotesingle s QAP for two-graph matching directly. Also, we show the extension to hypergraph matching, i.e. Neural Hyper-Graph Matching~(NHGM) in Sec.~\ref{sec:nhgm}, and to multiple graph matching i.e. Neural Multi-Graph Matching~(NMGM) in Sec.~\ref{sec:nmgm}. All these three settings to our knowledge have been hardly addressed by neural network solvers before. In Sec.~\ref{sec:ngmv2} we introduce NGM/NHGM/NMGM-v2 with significantly improved image matching accuracy by exploiting enhanced feature extractor.

%In Sec.~\ref{sec:ngm+} the enhanced model NGM+ is devised by introducing edge embeddings.
%Our approaches are not only able to learn scoring and embedding networks from the input affinity matrix/tensor, which is the general input for QAP (or even higher-order case), but also capable of handling image inputs (as \cite{ZanfirCVPR18,WangICCV19}), where an additional CNN is typically learned as a keypoint feature extractor for matching images. In fact, none of the existing graph matching deep learning methods can handle the first case i.e. dealing with Lawler\textquotesingle s QAP directly. The overview of our approach is shown in Fig.~\ref{fig:overview}.

%\subsection{Preliminaries}

%\subsubsection{Notations}

\subsection{Preliminaries}\label{sec:preliminary}
Our models aim to match weighted graph ${G}^1=({V}^1, {E}^1)$ and ${G}^2=({V}^2, {E}^2)$ (in capital letters), where the superscript means the index of graphs and the subscript means the index of nodes. Without loss of generality, $|{V}^1| = n_1 = n_{in}$ are all inlier nodes, and $|{V}^2| = n_2 = n_{in} + n_{out}$ contains both inliers and optional outliers. ${E}^1, {E}^2$ are attributed edge sets with second-order features in graphs and $|E^1|=n_{e1}, |E^2|=n_{e2}$. Lawler\textquotesingle s QAP as given in Eq.~(\ref{eq:lawler_qap}) is relaxed via popular doubly-stochastic relaxation:
\begin{align}\label{eq:convex_lawler_qap}
&J(\mathbf{S}) = \text{vec}(\mathbf{S})^\top\mathbf{K}\text{vec}(\mathbf{S}),\\\notag
&\mathbf{S}\in [0,1]^{n_1\times n_2}, \quad\mathbf{S}\mathbf{1}_{n_2}= \mathbf{1}_{n_1}, \quad \mathbf{S}^\top\mathbf{1}_{n_1}\leq\mathbf{1}_{n_2}
\end{align}
where $\mathbf{S}$ is a (partial) doubly-stochastic matrix, where all its rows sum to 1 and all its column sums are $\leq 1$. For affinity matrix $\mathbf{K} \in \mathbb{R}^{n_1n_2 \times n_1n_2}$, diagonal elements $\mathbf{K}_{ia,ia} = \mathbf{s}_v(V^1_i, V^2_a)$ are first order (node) similarities and off-diagonal elements $\mathbf{K}_{ia,jb}=\mathbf{s}_e(E^1_{ij}, E^2_{ab})$ are second order (edge) similarities, where $\mathbf{s}_v, \mathbf{s}_e$ are similarity measurements for nodes and edges, respectively. %Our proposed network learns a matching $\mathbf{X}$ from the patterns of $\mathbf{K}$ which empirically optimizes $J(\mathbf{X})$.
 
%\subsection{Problem Formulation}

%In the following, we denote input graphs by capitals $G = (V, E)$, and consider the matching problem between $G^1 = (V^1, E^1)$ and $G^2 = (V^2, E^2)$, where $|V^1|=n_1, |E^1|=n_{e1}, |V^2|=n_2, |E^2|=n_{e2}$. The superscript means the index of graphs and the subscript represents the index of nodes. 
As shown in Fig.~\ref{fig:association_graph}, graph matching can be viewed in a perspective based on the definition of the so-called association graph $\mathcal{G}^A=(\mathcal{V}^A, \mathcal{E}^A)$~\cite{ChoECCV10,LeordeanuICCV05} (in handwritten letters with superscript $A$). \textbf{To avoid ambiguity between graphs and association graphs, we name the entities in graphs ($V$) as \emph{nodes} and the entities in association graph ($\mathcal{V}^A$) as \emph{vertices}.} Readers should distinguish these two concepts as they will be repeatedly encountered through this paper. 

The vertices of association graph $\mathcal{V}^A = V^1 \times V^2$ encode candidate node-to-node correspondence $\mathcal{V}^A_{ia} = (V^1_i, V^2_a)$ corresponding to the matching matrix $\mathbf{X}_{i,a}$, therefore the vectorized assignment matrix $\text{vec}(\mathbf{X})$ is equivalent to the vertex set of association graph. The edges $\mathcal{E}^A$ represent the agreement between two pairs of correspondence $\mathcal{E}^A_{ia,jb}=\{(V^1_i, V^2_a), (V^1_j, V^2_b)\}$ modeled by $\mathbf{K}_{ia,jb}$, so that the off-diagonal part of affinity matrix $\mathbf{K}$ is equivalent to the adjacency matrix of association graph. The matching between two graphs can therefore be transformed into vertex classification on the association graph, following \cite{ChoECCV10,LeordeanuICCV05}. In this paper, diagonal elements $\mathbf{K}_{ia,ia}$ are further assigned as vertex attributes $\mathcal{V}^A$, to better exploit the first-order similarities. Such formulation can also be generalized to hypergraph matching problems, with edges replaced by hyperedges, as shown in Fig.~\ref{fig:association_graph}(c). The association graph prohibits links that violate the one-to-one matching constraint (e.g.\ there is no link between vertex `1a' and `1c' in Fig.~\ref{fig:association_graph}(a)).

\subsection{NGM: Neural Graph Matching for QAP}
\label{sec:ngm}
%\subsubsection{Overview}
Based on the formulations and association graph introduced in Sec.~\ref{sec:preliminary}, our proposed Neural Graph Matching (NGM) solves relaxed Lawler\textquotesingle s QAP in Eq.~(\ref{eq:convex_lawler_qap}), by vertex classification via Graph Convolutional Networks (GCN)~\cite{KipfICLR17} with novel matching-aware embedding modules. The vertex classification is performed on the association graph induced by the affinity matrix, followed by a Sinkhorn operator. As shown by existing learning-free graph matching solvers~\cite{ChoECCV10,LeordeanuICCV05}, graph matching problem is equivalent to vertex classification on the association graph. NGM accepts either raw image (with jointly learned CNN and affinity metric), or affinity matrix (without CNN or affinity metric), and learns end-to-end from ground truth correspondence or by pure optimization for QAPLIB problems. 
%We omit the details of this part to avoid distraction, and readers are referred to \cite{ZanfirCVPR18} for details. In the following, we describe the main steps of NGM.

\begin{figure}[tb!]
    \centering
    \includegraphics[width=0.35\textwidth]{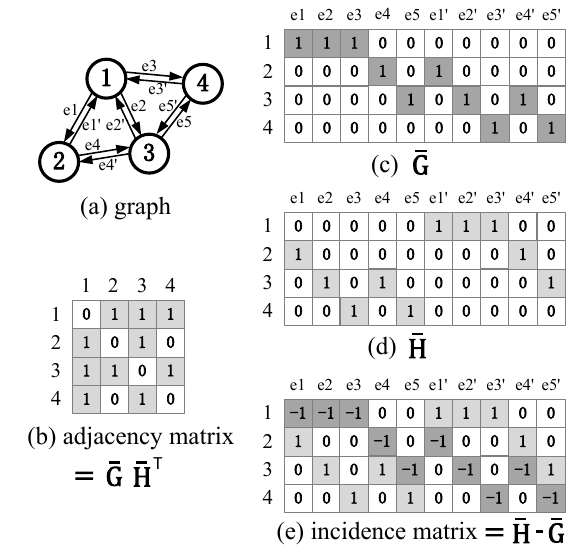}
    %\vspace{-10pt}
    \caption{A toy example of connectivity matrices (c) $\bar{\mathbf{G}}$, (d) $\bar{\mathbf{H}}$, and their connection to (a) the original graph, (b) adjacency matrix and (e) incidence matrix. All edges are directed, and ${\bar{\mathbf{G}}}_{i,k}={\bar{\mathbf{H}}}_{j,k}=1$ means edge $k$ starts from node $i$ and ends at node $j$. In incident matrix, -1 denotes the starting node and 1 denotes the ending node of all edges.}
    \label{fig:illustration_GH}
\end{figure}

\begin{figure*}[tb!]
    \centering
    \includegraphics[width=.9\textwidth]{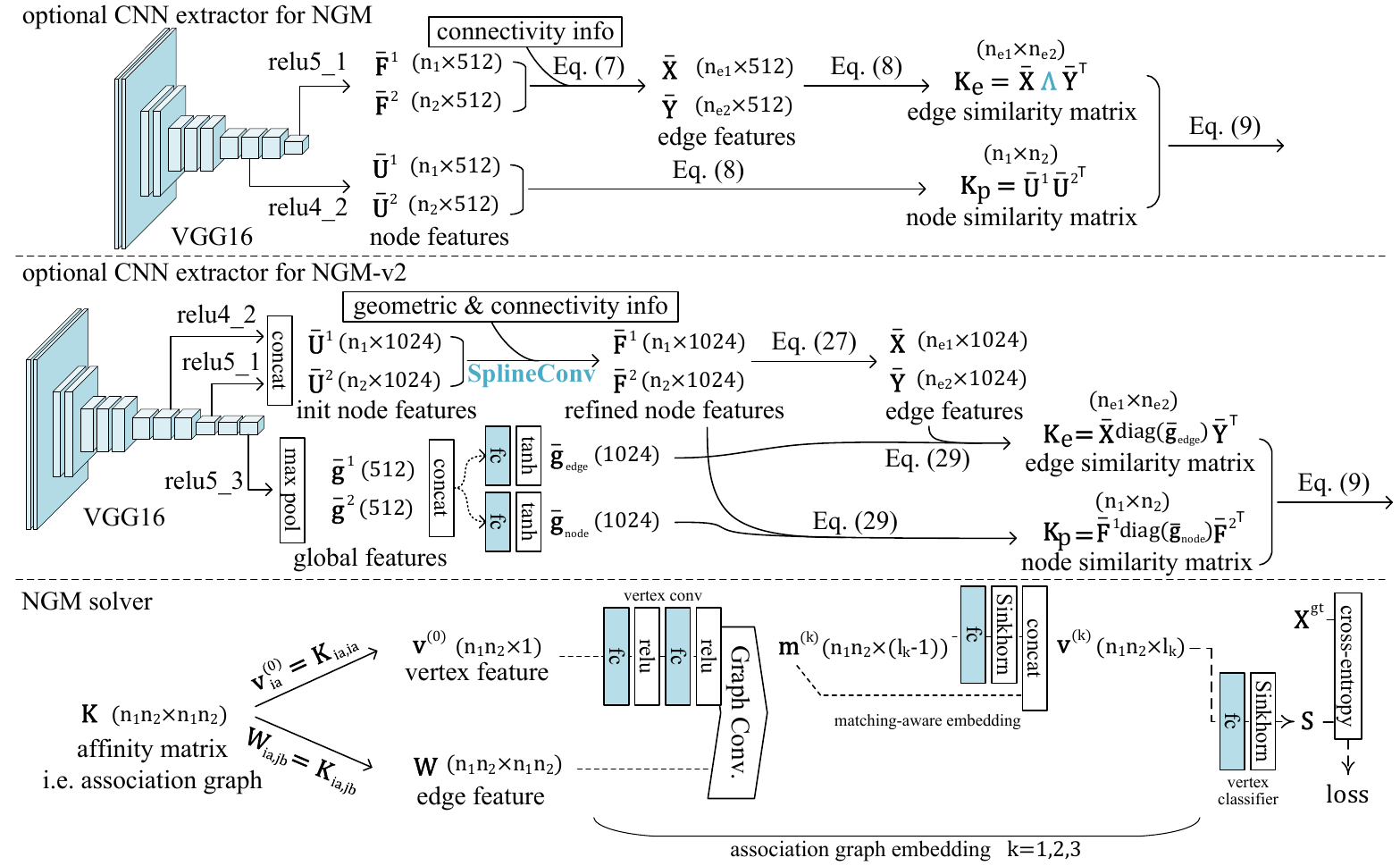}
     %\vspace{-10pt}
    \caption{The proposed NGM \& NGM-v2 architecture for two-graph matching. The components with blue FC layers i.e. vertex convolution, matching-aware embedding module and vertex classifier are jointly learned with optional CNN, SplineConv (in NGM-v2) and similarity metric $\mathbf{\Lambda}$ (in NGM).}
    \label{fig:network_structure}
\end{figure*}

\subsubsection{Affinity matrix building from natural images}\label{sec:model_CNN4imgs}
Our graph matching net allows either affinity matrix or raw images as input. The image processing module is optional and treated as a plug-in for dealing with images, following the protocol in \cite{ZanfirCVPR18} whereby the affinity matrix is built from pre-given keypoints in images. As shown in the upper half of Fig.~\ref{fig:network_structure}, image features are extracted by learnable CNN layers such as VGG16~\cite{simonyanICLR14vgg}. Given two input images with labeled keypoints, we adopt CNN layers to extract per-node features $\bar{\mathbf{F}}^1, \bar{\mathbf{U}}^1 \in \mathbb{R}^{n_1\times d}$ for $G^1$ and $\bar{\mathbf{F}}^2, \bar{\mathbf{U}}^2 \in \mathbb{R}^{n_2\times d}$ for $G^2$, where $d$ is feature dimension size and $\bar{\mathbf{F}}, \bar{\mathbf{U}}$ are extracted from different CNN layers (e.g.\ VGG16 \texttt{relu5\_1} for $\bar{\mathbf{F}}$ and \texttt{relu4\_2} for $\bar{\mathbf{U}}$) and utilized for edge representation and node representation, respectively. Features are obtained by bi-linear interpolation on the CNN feature map. As shown in Fig.~\ref{fig:illustration_GH}, the connectivity of two graphs are represented by $\bar{\mathbf{G}}^1,\bar{\mathbf{H}}^1\in \{0,1\}^{n_1\times n_{e1}}$ and $\bar{\mathbf{G}}^2,\bar{\mathbf{H}}^2\in \{0,1\}^{n_2\times n_{e2}}$, where $\bar{\mathbf{A}}^1 = \bar{\mathbf{G}}^1\bar{\mathbf{H}}^{1\top}, \bar{\mathbf{A}}^2 = \bar{\mathbf{G}}^2\bar{\mathbf{H}}^{2\top}$ are the adjacency matrices of two graphs, and ${\bar{\mathbf{G}}}_{i,k}={\bar{\mathbf{H}}}_{j,k}=1$ means edge $k$ links node $i$ to node $j$. The edge representations are built by concatenating node features at both ends of the edge:
\begin{equation}
    \bar{\mathbf{X}}=[\bar{\mathbf{G}}_1^\top \bar{\mathbf{F}}_1 \quad \bar{\mathbf{H}}_1^\top \bar{\mathbf{F}}_1], \ \bar{\mathbf{Y}}=[\bar{\mathbf{G}}_2^\top \bar{\mathbf{F}}_2 \quad \bar{\mathbf{H}}_2^\top \bar{\mathbf{F}}_2]
\end{equation}
where $[\ \cdot \quad \cdot \ ]$ means concatenating two matrices along columns. The node-to-node similarity matrix $\mathbf{K}_p\in \mathbb{R}^{n_1\times n_2}$ and edge-to-edge similarity $\mathbf{K}_e\in \mathbb{R}^{n_{e1}\times n_{e2}}$ are built via
\begin{equation}
    \mathbf{K}_e = \bar{\mathbf{X}} \mathbf{\Lambda} \bar{\mathbf{Y}}^\top, \mathbf{K}_p = \bar{\mathbf{U}}^1 \bar{\mathbf{U}}^{2\top}
\end{equation}
where $\mathbf{\Lambda}\in\mathbb{R}^{2d\times 2d}$ is the learnable parameter for affinity metric. The QAP affinity matrix is built following the factorized formulation of $\mathbf{K}$~\cite{ZhouPAMI16}:
\begin{equation}
    \mathbf{K}=\mathrm{diag}(\mathrm{vec}(\mathbf{K}_p)) + (\bar{\mathbf{G}}_2 \otimes_{\mathcal{K}} \bar{\mathbf{G}}_1) \mathrm{diag}(\mathrm{vec}(\mathbf{K}_e)) (\bar{\mathbf{H}}_2 \otimes_{\mathcal{K}} \bar{\mathbf{H}}_1)^\top
\label{eq:fgm}
\end{equation}
where $\mathrm{diag}(\cdot)$ means building a diagonal matrix from input vector, and $\otimes_{\mathcal{K}}$ means Kronecker product. All the forementioned operations allow back propagation, and we adopt the efficient GPU implementation provided by \cite{WangICCV19}. 

%The proposed network is based on GCN~\cite{KipfICLR17}, where node embeddings are aggregated through a fixed adjacency matrix. 
%Our GNN, in the meantime, also embeds edge features that are treated as adjacency matrices during aggregation. Such edge embedding can be found in \cite{gongCVPR19exploiting} exploiting edge features of graph networks. 
%(if there are any), following the seminal pipeline in \cite{ZanfirCVPR18} incorporating learnable weights in CNN and affinity mapping.

\subsubsection{Association graph construction}
We derive the association graph from the affinity matrix $\mathbf{K}$. The weighted adjacency matrix of association graph $\mathbf{W}$ comes from the off-diagonal elements of $\mathbf{K}$.
%On GNN layer $k$, we denote $\mathbf{W}^{(k)} \in \mathbb{R}^{n_1n_2\times n_1n_2 \times m_k}$ as $m_k$-dimensional edge embeddings, 
We denote $\mathbf{v}^{(k)} \in \mathbb{R}^{n_1n_2 \times l_k}$ as $l_k$-dimensional vertex embeddings on layer $k$ (starting with $k=0$). The initial embeddings at $k=0$ are scalar, i.e. $l_0=1$, taken from the diagonal of $\mathbf{K}$.
\begin{equation}
        \mathbf{W}_{ia,jb} = \mathbf{K}_{ia,jb},\quad \mathbf{v}^{(0)}_{ia} = \mathbf{K}_{ia,ia}
\end{equation}
$\mathbf{W}$ contains both connectivity and weight information in the association graph. In case when the first-order similarity $\mathbf{K}_{ia,ia}$ is absent, we can assign a constant (e.g.\ 1) for all $\mathbf{v}^{(0)}$.

\subsubsection{Matching aware embedding of association graph}
The matching problem can be transformed to selecting the vertices in the association graph that encode the node-to-node correspondence between two input graphs, as shown in Fig.~\ref{fig:association_graph}. For vertex classification on the association graph, we use GCN~\cite{KipfICLR17} for its effectiveness and simplicity.

We define the (unweighted) adjacency matrix of association graph $\mathbf{A} \in \{0,1\}^{n_1n_2 \times n_1n_2}$: $\mathbf{A}_{ia,jb} = 1$ if $\mathbf{K}_{ia,jb}>0$ and otherwise 0. $\mathbf A_{ia,jb}$ serves as an indicator whether there exists an edge between vertices $ia$ and $jb$ in the association graph. Recall that the vertex $ia$ represents the matching between node $i$ and $a$ from separate input graphs (see Fig.~\ref{fig:association_graph}). In association graph, an edge exists between $ia$ and $jb$ if and only if the node-to-node matchings $i$ to $a$ and $j$ to $b$ can co-exist, and there exists an edge-to-edge affinity score defined between edges $ij$ and $ab$. Since $\mathbf{A}$ is symmetric, we compute the degree matrix for normalization: 
\begin{equation}
    %\mathbf{A}^\prime = \mathbf{A} \oslash ( \mathbf{1} \mathbf{1}^\top \mathbf{A})
    \mathbf{D} = \mathrm{diag}(\mathbf{A} \mathbf{1}_{n_1n_2})
\end{equation}
where $\mathrm{diag}(\cdot)$ builds a diagonal matrix from input vector. The vertex aggregation step is according to:
\begin{equation}
%\begin{split}
\label{eq:node}
%\mathbf{m}^{(k)} = f_n(\mathbf{v}^{(k-1)}), \quad 
    %\mathbf{v}^{(k)} = \mathbf{P}^{(k)} \circledast_3 \mathbf{m}^{(k)}
    \mathbf{m}^{(k)} = \mathbf{D}^{-1}  \mathbf{W} f_m(\mathbf{v}^{(k-1)}) + f_v(\mathbf{v}^{k-1}), \
    \mathbf{v}^{(k)} = \mathbf{m}^{(k)}
    %&\mathbf{m}^{(k)} = f_n(\mathbf{v}^{(k-1)}), \quad \mathbf{n}^{(k)} = \mathbf{P}^{(k)} \circledast \mathbf{m}^{(k)},\\
    %&\mathbf{v}^{(k)} = \left[\mathbf{n}^{(k)} \quad \text{vec}(\text{SK}(\mathbf s(\mathbf{n}^{(k)})))\right]
%\end{split}
\end{equation}
where the message passing function $f_m: \mathbb{R}^{l_{k-1}}\rightarrow \mathbb{R}^{l_k}$ and vertex\textquotesingle s self update function $f_v: \mathbb{R}^{ l_{k-1}}\rightarrow \mathbb{R}^{l_k}$ are both implemented by networks with two fully-connected layers and ReLU activation.

The above general vanilla vertex embedding procedure in Eq.~(\ref{eq:node}) does not consider the one-to-one assignment constraint for matching. Here we develop a matching constraint aware embedding model: in each layer a soft permutation (i.e. doubly-stochastic matrix) is scored via classifier with Sinkhorn network $\mathrm{Classifier}: \mathbb{R}^{n_1n_2\times l_k}\rightarrow [0,1]^{n_1\times n_2}$ (see discussions in Sec.~\ref{sec:sinkhorn_classifier}) followed by vectorization operator $\mathrm{vec}(\cdot)$. The predicted soft permutation is concatenated to vertex embeddings whereby matching information is considered in embedding layers. 
With $f_m, f_v: \mathbb{R}^{l_{k-1}}\rightarrow \mathbb{R}^{(l_k-1)}$, such a matching-aware embedding scheme is denoted as \textbf{Sinkhorn embedding} in the rest of the paper.
\begin{equation}
\label{eq:sk_emb}
%\begin{split}
    %\mathbf{n}^{(k)} = \text{vec}(\text{Classifier}(\mathbf{m}^{(k)})), 
    \mathbf{v}^{(k)} = [\mathbf{m}^{(k)} \quad \mathrm{vec}(\mathrm{Classifier}(\mathbf{m}^{(k)}))] 
%\end{split}
\end{equation}
where $[\ \cdot \quad \cdot \ ]$ means concatenation. We experiment both vanilla vertex embedding in Eq.~(\ref{eq:node}) and matching-aware Sinkhorn embedding in Eq.~(\ref{eq:sk_emb}), to validate the necessity of adding assignment constraint.

%One GNN layer is constructed by concatenating an edge embedding step and a node embedding step.

\subsubsection{Vertex classification with Sinkhorn network}
\label{sec:sinkhorn_classifier}
As graph matching is equivalent to vertex classification on association graph (see Fig.~\ref{fig:association_graph}), a vertex classifier with Sinkhorn network is adopted to predict the matching result. Specifically we use a single layer fully-connected classifier denoted by $f_c: \mathbb{R}^{l_k}\rightarrow \mathbb{R}$, followed by exponential activation with regularization factor $\alpha$:
\begin{equation}
    \label{eq:sk_regularization}
    \mathbf{s}_{ia}^{(k)} = \exp \left(\alpha f_c(\mathbf{v}_{ia}^{(k)})\right)
\end{equation}

After reshaping classification scores into $\mathbb{R}^{n_1 \times n_2}$, one-to-one assignment constraint is enforced to $\mathbf{s}$ by Sinkhorn network~\cite{AdamsArxiv11,menaICLR18learning}. It takes a non-negative square matrix as input and outputs a doubly-stochastic matrix~\cite{AdamsArxiv11,SantaPAMI18}. As the scoring matrix can be non-square for different sizes of graphs, the input matrix $\mathbf{S} \in \mathbb{R}^{n_1 \times n_2}$ is padded into a square one (assume $n_1 \leq n_2$) with small elements e.g.\ $\epsilon = 10^{-3}$. A doubly-stochastic matrix is obtained by repeatedly running:
\begin{equation}
%\label{eq:sk_col}
    \mathbf{S} = \mathbf{S} \oslash (\mathbf{1}_{n_2} \mathbf{1}_{n_2}^\top \cdot \mathbf{S}), \
%\label{eq:sk_row}
    \mathbf{S} = \mathbf{S} \oslash (\mathbf{S} \cdot \mathbf{1}_{n_2} \mathbf{1}_{n_2}^\top)
    \label{eq:sk_rowcol}
\end{equation}
where $\oslash$ means element-wise division. By taking column-normalization and row-normalization in Eq.~(\ref{eq:sk_rowcol}) alternatively, $\mathbf{S}$ converges to a doubly-stochastic matrix whose rows and columns all sum to 1. The dummy elements are discarded in the final output, whose column sum may be $<1$ given umatched nodes from the bigger graph. Sinkhorn operator is fully differentiable and can be efficiently implemented by automatic differentiation techniques~\cite{serratosa2011automatic}. The proposed vertex classifier with Sinkhorn network is denoted as $\mathrm{Classifier}: \mathbb{R}^{n_1n_2\times l_k}\rightarrow [0,1]^{n_1\times n_2}$, which is also mentioned in matching-aware Sinkhorn embedding in Eq.~(\ref{eq:sk_emb}).

\subsubsection{Loss for end-to-end training}
Recall the obtained predicted matrix $\mathbf{S}$ from the above procedure is a doubly-stochastic matrix. Each element can be regarded as a binary classification where each vertex should be classified to 1 (matched) or 0 (unmatched). Hence we adopt the binary cross-entropy as the final loss, given the ground truth node-to-node correspondence $\mathbf{X}^{gt}$:
\begin{equation}
\label{eq:perm_loss}
    \ell=- \sum_{i = 1}^{n_1} \sum_{a = 1}^{n_2} \mathbf{X}^{gt}_{i,a} \log \mathbf{S}_{i,a} + (1-\mathbf{X}^{gt}_{i,a}) \log (1-\mathbf{S}_{i,a}) 
\end{equation}
Our approach also allows direct optimization over the affinity score objectives for QAPLIB problems, which will be discussed in Sec.~\ref{sec:exp_qaplib} in details.

All the components are differentiable. Therefore, both NGM solver and the optional CNN feature extractor can be learned via backpropagation and gradient descent. %We further enable NGM with edge embedding.

\subsection{NHGM: Neural Hypergraph Matching}
\label{sec:nhgm}
For Neural Hyper-Graph Matching (NHGM), higher-order structure is exploited for more robust matching. NHGM owns a nearly identical pipeline compared to NGM, while a more general message-passing scheme is devised for vertex classification in hypergraphs, as previously shown in \cite{bai2021hypergraph,FengAAAI19}. Due to the explosive computational cost ($O((n_1n_2)^t)$ with order $t$), here we limit hypergraph to third-order which is also in line with the majority of existing works~\cite{YanCVPR15,LeeCVPR11}, while the scheme is generalizable to any order $t$.

%\textbf{Modeling higher-order affinity}.%, such as the third-order one $\mathbf{H} \in \mathbb{R}^{n_1n_2\times n_1n_2 \times n_1n_2}$. 
%Hypergraphs are composed of nodes and hyperedges, where two or more nodes may be linked by each hyperedge.

\begin{figure}
    \centering
    \includegraphics[width=0.3\textwidth]{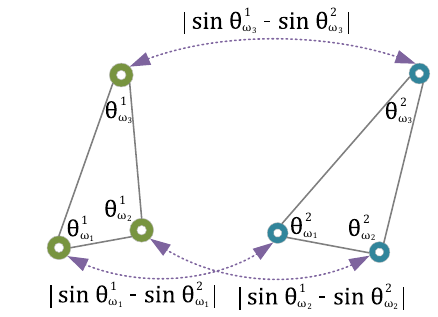}
     %\vspace{-10pt}
    \caption{Third-order affinity adopted in our hypergraph matching. It in general follows the previous hypergraph matching works~\cite{YanCVPR15,LeeCVPR11}, by considering the similarity between two triangle\textquotesingle s three inner angles.}
    \label{fig:3rd_affinity}
\end{figure}

\subsubsection{Association hypergraph construction}
The second-order affinity matrix $\mathbf{K}$ is generalized to affinity tensor $\mathbf{H}^{\left<t\right>}$ of order $t$ in hypergraph matching. In line with the hypergraph matching literature~\cite{ZassCVPR08,LeeCVPR11,DuchennePAMI11}, the third-order affinity tensor is specified as:
\begin{equation}
\label{eq:3order_aff}
    \mathbf{H}^{\left<3\right>}_{\omega_1, \omega_2, \omega_3} = \exp \left( - \left(\sum_{q=1}^{3} |\sin \theta^1_{\omega_q} - \sin \theta^2_{\omega_q}|\right)/\sigma_3\right)
\end{equation}
where $\theta^1_{\omega_q}, \theta^2_{\omega_q}$ denotes the angle in graph $\mathcal{G}_1$ and $\mathcal{G}_2$ of each correspondence $\omega_q$, respectively. An illustration of third order affinity can be found in Fig.~\ref{fig:3rd_affinity}, where the similarity between triplets of nodes is compared. Third order affinity is usually defined on geometric consistency and it preserves both scaling and rotation invariance.
%Readers are referred to Fig.~5 in \cite{LeeCVPR11} for an intuitive view of the third-order affinity. 
%Lower order affinities are also encoded into the affinity tensor, regarding the \emph{learnable} weights $\{\lambda_t\}$:
%\begin{equation}
%\label{eq:aff_tensor}
%    \textbf{H}^{\left<t\right>}_{\omega_1, \cdots, \omega_t} = \Omega^{\left<t\right>}_{\omega_1, \cdots, \omega_t} + \lambda_{t-1} \sum_{p=1}^t  \textbf{H}^{\left<t - 1\right>}_{\omega_1, \cdots, \omega_t / \omega_p}
%\end{equation}

Extending from the second-order association graph, a hyper association graph is built from $\mathbf{H}$. The association hypergraph $\mathcal{H}^{A} = (\mathcal V^{A}, \mathcal E^{A})$ takes node-to-node correspondence $\omega = (V^1_i, V^2_j)$ as vertices $\mathcal V^{A}$ (which is consistent with second-order association graph) and higher-order similarity among $\{(V^1_{\omega_1}, V^2_{\omega_1}), \cdots, (V^1_{\omega_t}, V^2_{\omega_t})\}$ as hyperedges $\mathcal E^{A}$, as shown by Fig.~\ref{fig:association_graph}(c).  Elements of $\mathbf{H}$ are adjacency weights for the association hypergraph accordingly. In NHGM, hypergraph convolution is defined for vertex classification.

%\textbf{Hyperedge embedding}. In line with edge embedding of second order graphs, we propose hyperedge embedding with a more generalized feature update scheme for hyperedges. Firstly, we define the unweighted hyper adjacency tensor: $\mathbf{A}_{\omega_1,\cdots,\omega_k} = 1$, if $\mathbf{H}_{\omega_1,\cdots,\omega_k}>0$ and otherwise $0$.
%\begin{equation}
 %   \mathbf{A}_{\omega_1,\cdots,\omega_k} = \left\{ 
  %  \begin{aligned}
   % &1, \quad \mathbf{H}_{\omega_1,\cdots,\omega_k}>0 \\
    %&0, \quad otherwise
    %\end{aligned}
    %\right.
%\end{equation}

%Edge embeddings are initialized as $\textbf W^{(0)} = \textbf H$, and updated through neural network $f_e(\cdot)$:
%\begin{align}
%    \mathbf{W}^{(k)}_{\omega_1, \cdots, \omega_t} = f_e\left(\left[\mathbf{W}^{(k-1)}_{\omega_1, \cdots, \omega_t} \quad \sum_{p=2}^t \mathbf{v}^{(k-1)}_{\omega_p}\right]\right) \cdot \mathbf{A}_{\omega_1, \cdots, \omega_t} %+ \sum_{p=2}^t f_{n\rightarrow e}(\mathbf{v}^{(k-1)}_{\omega_p}) \cdot \mathbf{A}_{\omega_1, \cdots, \omega_t}
%\end{align}
%Edge features are aggregated from itself and its adjacent nodes (except the node on the first dimension of adjacency tensor, as it will be updated by this feature in the next node embedding step).

\subsubsection{Matching aware association hypergraph embedding}
As an extension of Eq.~(\ref{eq:node}), vertex embeddings are updated from all vertices linked by hyperedges in association hypergraph. We compute normalized degree tensor of order $t$:
\begin{equation}
    {\mathbf{D}^{\left<t\right>\ -1}_{\omega_1, \cdots, \omega_t}} = \mathbf{A}^{\left<t\right>}_{\omega_1, \cdots, \omega_t} / (\mathbf{A}^{\left<t\right>} \otimes_2 \textbf 1 \cdots \otimes_t \textbf{1})_{\omega_1} 
    %\mathbf{P}^{(k)} &= \mathbf{W}^{(k)} \odot_{t+1} \mathbf{A}^\prime
\end{equation}
Then an aggregation scheme extended from Eq.~(\ref {eq:node}) is taken:
\begin{equation}
\begin{split}
    \mathbf{p}^{(k)} &= f_m^{\left<t\right>}(\mathbf{v}^{(k-1)}), \ \mathbf{H}^{\left<t\right>\prime} = (\mathbf{D}^{\left<t\right>\ -1} \odot \mathbf{H}^{\left<t\right>})_{t+1} \\
    \mathbf{m}^{(k)} &= \sum_t \lambda_t \ \mathbf{H}^{\left<t\right>\prime} \otimes_t \mathbf{p}^{(k)} \cdots \otimes_2 \mathbf{p}^{(k)} + f_v \\
    \mathbf{v}^{(k)} &= \left[\mathbf{m}^{(k)} \quad \mathrm{vec}(\mathrm{Classifier}(\mathbf{m}^{(k)}))\right]
\end{split}
\end{equation}
where $f_v$ abbreviates $f_v(\mathbf{v}^{(k-1)})$, $\otimes_i$ denotes tensor product by dimension $i$ (see Eq.~(\ref{eq:tensor_prod})), $\odot$ denotes element-wise multiply, $(\cdot)_{t+1}$ means expanding along dimension $(t+1)$. $f_m^{\left<t\right>}:\mathbb{R}^{l_{k-1}}\rightarrow \mathbb{R}^{(l_k-1)}$ is message passing function at order $t$. Different orders of features are fused by weighted summation with $\lambda_t$. 

The other modules of NHGM, including classifier and cross-entropy loss, are identical to NGM. Therefore, NGM can be viewed as a special case of NHGM, where the order is restricted to 2. The sparsity of $\mathbf{H}$ is exploited for both time and memory efficiency. 
%Note that the edge embedding scheme introduced in Sec.~\ref{sec:ngm+} is also applicable to NHGM, but we stick to the simple version for cost-effectiveness.

\subsection{NMGM: Neural Multi-graph Matching}
\label{sec:nmgm}
%, where matching information is fused among multiple graphs by so-called cycle-consistency
We explore learning multi-graph matching, where the information is fused among graphs by the so-called cycle-consistency. Cycle-consistency denotes a condition where the matching between any two graphs is consistent when passed through any other graphs, i.e. $\mathbf{X}_{ij}=\mathbf{X}_{ik}\mathbf{X}_{kj}$ for all $i, j, k$, which can be viewed as each graph matched to a $n$-sized reference ${\mathbf X}_i\in\{0,1\}^{n\times n}$. The pariwise matching between $G_i,G_j$ can be represented with ${\mathbf{X}}_{ij} = {\mathbf{X}}_i {\mathbf{X}}_j^\top$. In this paper, we refer to the line of works~\cite{PachauriNIPS13,chen2014near,ZhouICCV15,MasetICCV17} involving post-synchronization given initial pairwise matchings. Spectral fusion is used in NMGM for its effectiveness and simplicity, and most importantly, its ability for end-to-end training. We assume all graphs are of equal size. 

%\subsubsection{Generalizing multi-matching for end-to-end}
%Existing spectral multi-matching algorithms work with discrete permutation matrices, which are, however, not capable for gradient propagation. In this paper, such constraints are relaxed to doubly stochastic matrices for their compatibility with end-to-end learning. For each pair ${G}_i$ and ${G}_j$ with $n$ nodes, $\mathbf{S}_{ij}\in[0,1]^{n\times n}$ is computed by NGM as the soft (i.e. doubly-stochastic) matching matrix. Here we generalize the definition of cycle-consistency from permutation matrix to doubly-stochastic matrix: $\hat{\mathbf{S}}_{ij}=\hat{\mathbf{S}}_{ik}\hat{\mathbf{S}}_{kj}$ for all $i, j, k$. A cycle-consistent matching can be regarded as each graph matched to a $n$-sized \textit{universe} $\hat{\mathbf S}_i\in[0,1]^{n\times n}$ and pariwise matching between $G_i,G_j$ can be represented with $\hat{\mathbf{S}}_{ij} = \hat{\mathbf{S}}_i \hat{\mathbf{S}}_j^\top$.

\subsubsection{Building joint matching matrix}
We first obtain initial two-graph matchings by NGM to build a symmetric joint matching matrix $\mathcal{S}\in \mathcal{R}^{nm\times nm}$. For each pair ${G}_i$ and ${G}_j$ with $n$ nodes, $\mathbf{S}_{ij}\in[0,1]^{n\times n}$ is computed by NGM as the soft (i.e. doubly-stochastic) matching matrix. 
For $m$ graphs, $\mathcal{S}$ can be built from all combinations of pairwise matchings $\mathbf{S}_{ij}$:
\begin{equation}
\label{eq:joint_matching}
\mathcal{S} = \left(
\begin{array}{ccc}
    \mathbf{S}_{00} & \cdots & \mathbf{S}_{0m} \\
    \vdots & \ddots & \vdots \\
    \mathbf{S}_{m0} & \cdots & \mathbf{S}_{mm}
\end{array}
\right)
\end{equation}
where $\mathcal{S}$ is of size $mn\times mn$. For the diagonal part of $\mathcal{S}$, $\mathbf{S}_{ii}$ are all identical matrices. Note $\mathbf{S}_{ij}$ are all square matrices. The objective of spectral fusion results in getting a cycle-consistent joint matching matrix $\hat{\mathcal{S}}$, whose innerproduct with $\mathcal{S}$ is maximized:
\begin{equation}
    \label{eq:mgm_obj}
    \max_{\hat{\mathcal{S}}} = \mathrm{tr}(\hat{\mathcal{S}}^\top \mathcal{S})
\end{equation}
To ensure the cycle-consistency, it holds $\hat{\mathbf{S}}_{ij} = \hat{\mathbf{S}}_{ik} \hat{\mathbf{S}}_{kj}$ for all elements in $\hat{\mathcal{S}}$. We may select an arbitrary $k$ as the reference, omitting $k$ in the subscript, $\hat{\mathcal{S}}$ can be decomposed as matching to the reference:
\begin{equation}
\begin{split}
    \hat{\mathbf{U}} \hat{\mathbf{U}}^\top &= \hat{\mathcal{S}} \\
     \text{where} \quad \hat{\mathbf{U}} &= \left(\begin{array}{c}
        \hat{\mathbf{S}}_0 \\
        \vdots \\
        \hat{\mathbf{S}}_m
    \end{array}\right)
\end{split}
\end{equation}
%each column of $\hat{\mathbf{U}}$ is linearly independent unless all elements equal to $1/n$ (which can be prohibited by setting proper Sinkhorn regularization parameter $\alpha$ in NGM). Therefore, $\hat{\mathbf{U}}/\sqrt{m}$ are $n$ eigenvectors of $\hat{\mathcal{S}}$ with eigenvalue $m$. Eq.~(\ref{eq:mgm_obj}) is a generalized Rayleigh problem and solved via spectral fusion, as shown follows.
under ideal condition where $\hat{\mathbf{S}}_i$ are all permutation matrices, each column of $\hat{\mathbf{U}}$ is linearly independent and $\hat{\mathbf{U}}/\sqrt{m}$ are $n$ eigenvectors of $\hat{\mathcal{S}}$ with eigenvalue $m$. The permutation constraint in $\hat{\mathbf{S}}_i$ is relaxed for computational feasibility and Eq.~(\ref{eq:mgm_obj}) results in a generalized Rayleigh problem and solved via spectral fusion, as shown follows. %The gap between theoretical condition can be controlled by setting proper Sinkhorn regularization parameter $\alpha$ in NGM. 

\subsubsection{Differentiable spectral fusion of pairwise matchings}
Multi-graph matching information can be fused by eigenvector decomposition (i.e. spectral method) on $\mathcal{S}$. Based on generalized Rayleigh problem, given $n$ nodes in each graph, we extract the eigenvectors corresponding to the top-$n$ eigenvalues of symmetric matrix $\mathcal{S}$:
\begin{equation}
    \label{eq:ed}
    \mathbf{U} \ \mathbf{\Sigma} \ \mathbf{U}^\top = \mathcal{S}
\end{equation}
where diagonal matrix $\mathbf{\Sigma} \in \mathbb{R}^{n\times n}$ contains top-$n$ eigenvalues and $\mathbf{U}\in \mathbb{R}^{mn \times n}$ are the $n$ corresponding eigenvectors. It has been shown that the computation of eigenvalues and eigenvectors are differentiable~\cite{IonescuICCV15} which makes them fixed components in our end-to-end learning network pipeline. The fusion of the input $\mathcal{S}$ can be written as follows, which can be seen as a smoothed version maximizing $\mathrm{tr}(\hat{\mathcal{S}}^\top \mathcal{S})$:
\begin{equation}
    \hat{\mathcal{S}} = m\ \mathbf{U}\ \mathbf{U}^\top
\end{equation}
%from which the synchronized two-graph matching matrices $\hat{\mathbf{S}}_{ij}$ can be obtained. 
The gradient of eigen decomposition in Eq.~(\ref{eq:ed}) is~\cite{IonescuICCV15}:
%\begin{equation}
\begin{align}
    \notag
    \frac{\partial L}{\partial {\mathcal{S}}} %&= \mathbf{U}\left(2\left(\mathbf{Y}^\top \odot \left(\mathbf{U}^\top \frac{\partial L}{\partial \hat{\mathcal{S}}} \frac{\partial \hat{\mathcal{S}}}{\partial \mathbf{U}}\right)_{sym}\right)
    %+\left(\frac{\partial L}{\partial \mathbf{\Sigma}}\right)_{diag}
    %\right) \mathbf{U}^\top\\
    &= \mathbf{U}\left(4m\left(\mathbf{Y}^\top \odot \left(\mathbf{U}^\top \left(\frac{\partial L}{\partial \hat{\mathcal{S}}} \right)_{sym} \mathbf{U}\right)_{sym}\right)
    %+\left(\frac{\partial L}{\partial \mathbf{\Sigma}}\right)_{diag}
    \right) \mathbf{U}^\top\\
   &\text{where}\qquad \mathbf{Y}_{ij} = \left\{ \begin{array}{ll}
        1 / (\sigma_i - \sigma_j) &i \neq j \\
        0 &i=j
            \end{array}\right. 
\end{align}
%\end{equation}
where $L$ denotes the loss, $\odot$ means element-wise multiplication, $\mathbf{A}_{sym}=(\mathbf{A}+\mathbf{A}^{\top})/2$
%, $\mathbf{A}_{diag}$ means setting all off-diagonal elements in $\mathbf{A}$ to 0 
and $\sigma_i=\mathbf{\Sigma}_{ii}$ is the $i$-th eigenvalue. 
%$\mathbf{\Sigma}$ is not involved in loss computation therefore ${\partial L}/{\partial \mathbf{\Sigma}}=0$.
According to this backward formulation, if there exist non-distinctive eigenvalues, i.e. $\sigma_i=\sigma_j \text{ for } i\neq j$, a numerical divided-by-zero error will be caused. This issue usually happens when cycle-consistency (i.e. $\mathbf{S}_{ij}=\mathbf{S}_{ik}\mathbf{S}_{kj}$) is already met in the input $\mathbf{S}$, under such circumstances the fused matching results are nearly identical to the original matchings. To avoid numerical issues, we assign $\hat{\mathbf{S}}_{ij} = \mathbf{S}_{ij}$ to bypass eigendecomposition if the minimum residual among top-$n$ eigenvalues is smaller than tolerance $\delta$, e.g.\ ${10}^{-4}$. This strategy is found effective to stabilize learning.

The final matching results are obtained by differentiable Sinkhorn network:
\begin{equation}
    \Bar{\mathbf{S}}_{ij} = \mathrm{Sinkhorn}(\exp(\hat \alpha \ \hat{\mathbf{S}}_{ij}))
\end{equation}
where the fused two-graph matching $\hat{\mathbf{S}}_{ij}$ is from $\hat{\mathcal{S}}$ and $\exp(\hat \alpha \ \cdot \ )$ performs regularization for Sinkhron. Cross-entropy loss in Eq.~(\ref{eq:perm_loss}) is applied to each $\Bar{\mathbf{S}}_{ij}$ for supervised learning, which is similar to the supervised two-graph matching case in the paper.

\subsection{Improved Matching by Enhanced Feature Extractor}
\label{sec:ngmv2}
Since our proposed method deals with the most general Lawler\textquotesingle s QAP, the baseline feature extractor in Sec.~\ref{sec:model_CNN4imgs} can be replaced by other enhanced feature extraction techniques. 
In this section, we refer to the novel feature extractor proposed by \cite{RolinekECCV20} in replacement of the feature extractor in Sec.~\ref{sec:model_CNN4imgs}, introducing the family of enhanced models for image matching problems -- NGM/NHGM/NMGM-v2. In Sec.~\ref{sec:v2_feature_extractor} we discuss how to build enhanced affinity matrix, and in Sec.~\ref{sec:v2_hypergraph} we propose a way of building hypergraph affinity tensor for NHGM-v2.

\subsubsection{Enhanced graph matching feature}
\label{sec:v2_feature_extractor}
The authors of \cite{RolinekECCV20} propose an enhanced deep learning feature extractor for graph matching problem on images based on SplineConv~\cite{FeyCVPR18} and weighted inner product, which can seamlessly fit into our pipeline. This enhanced feature extractor is summarized in the second row of Fig.~\ref{fig:network_structure}.

With some abuse of the notations from Sec.~\ref{sec:model_CNN4imgs}, we introduce the feature extractor as follows. Given two input images, we extract node features from \texttt{relu4\_2} and \texttt{relu5\_1} of VGG16, and then concatenate them to form the feature matrices $\bar{\mathbf{U}}^1 \in \mathbb{R}^{n_1 \times d}, \bar{\mathbf{U}}^2 \in \mathbb{R}^{n_2 \times d}$, where $d$ is the feature dimension of the concatenated features. Then we adopt two SplineConv\footnote{SplineConv is called SplineCNN in its original paper \cite{FeyCVPR18}. To avoid the ambiguity of the term ``CNN'' which usually represents convolution on images while SplineCNN is convolution on graphs, we name it as SplineConv in this paper.}~\cite{FeyCVPR18} layers on $\bar{\mathbf{U}}^1, \bar{\mathbf{U}}^2$ to produce refined features $\bar{\mathbf{F}}^1 \in \mathbb{R}^{n_1 \times d}, \bar{\mathbf{F}}^2 \in \mathbb{R}^{n_2 \times d}$. SplineConv~\cite{FeyCVPR18} is a powerful graph convolution operator exploiting B-Spline kernel, encoding geometric features into node features. Therefore, SplineConv is suitable for feature refinement on image matching datasets. Readers are referred to the original paper for details about SplineConv. The edge features are constructed as the difference of its two nodes in the feature space:
\begin{equation}
    \bar{\mathbf{X}}=\bar{\mathbf{G}}_1^\top \bar{\mathbf{F}}_1 - \bar{\mathbf{H}}_1^\top \bar{\mathbf{F}}_1, \ \bar{\mathbf{Y}}=\bar{\mathbf{G}}_2^\top \bar{\mathbf{F}}_2 - \bar{\mathbf{H}}_2^\top \bar{\mathbf{F}}_2
\end{equation}
where $\bar{\mathbf{G}}, \bar{\mathbf{H}}$ are the connectivity matrices in Fig.~\ref{fig:illustration_GH}.

As shown in the second row of Fig.~\ref{fig:network_structure}, the enhanced feature extractor also contains a branch producing global features by max-pooling over the output of \texttt{relu5\_3} layer. The pooled global features from two graphs are then concatenated, and passed to a \texttt{fc} layer followed with \texttt{tanh} activation, producing global features $\bar{\mathbf{g}}\in\mathbb{R}^d$:
\begin{equation}
    \bar{\mathbf{g}} = \text{tanh}(\text{fc}([\bar{\mathbf{g}}^1 \quad \bar{\mathbf{g}}^2]))
\end{equation}
where $[\ \cdot \quad \cdot \ ]$ means concatenation, and we compute separate $\bar{\mathbf{g}}_{\text{node}}, \bar{\mathbf{g}}_{\text{edge}}$ for node and edge features, respectively.

Node and edge similarity matrices are constructed based on weighted inner-product whereby $\bar{\mathbf{g}}$ as the weight:
\begin{equation}
    \mathbf{K}_e = \bar{\mathbf{X}} \  \mathrm{diag}(\bar{\mathbf{g}}_{\text{node}}) \ \bar{\mathbf{Y}}^\top, \mathbf{K}_p = \bar{\mathbf{F}}^1 \  \mathrm{diag}(\bar{\mathbf{g}}_{\text{edge}}) \ \bar{\mathbf{F}}^{2\top}
\end{equation}
where $\mathrm{diag}(\bar{\mathbf{g}})$ means building a diagonal matrix by placing $\bar{\mathbf{g}}$ on it. After obtaining $\mathbf{K}_e, \mathbf{K}_p$, the affinity matrix can be built by Eq.~(\ref{eq:fgm}), and the resulting QAP is readily solved by NGM/NMGM as discussed in Sec.~\ref{sec:ngm} and Sec.~\ref{sec:nmgm}.

\subsubsection{Enhanced hypergraph affinity for NHGM-v2}
\label{sec:v2_hypergraph}
Based on the the features defined in Sec.~\ref{sec:v2_feature_extractor}, we further design the hyperedge affinity for NHGM-v2. The hypergraph affinity is inspired by Fig.~\ref{fig:3rd_affinity}, however in NHGM the hypergraph affinity is based on geometric features, and in NHGM-v2 we define hypergraph affinity on the high-dimensional feature space. We regard the plane in Fig.~\ref{fig:3rd_affinity} as the feature space formed by $\bar{\mathbf{F}}^1$ and $\bar{\mathbf{F}}^2$, and each point in Fig.~\ref{fig:3rd_affinity} represents a node with features. Following Eq.~(\ref{eq:3order_aff}), the third order affinity tensor is defined as the differences of angles in the feature space:
\begin{equation}
\label{eq:nhgmv2_3order_aff}
    \mathbf{H}^{\left<3\right>}_{\omega_1, \omega_2, \omega_3} = \exp \left( - \left(\sum_{q=1}^{3} |\cos \theta^1_{\omega_q} - \cos \theta^2_{\omega_q}|\right)/\sigma_3\right)
\end{equation}
and we empirically find $\cos$ performs better than $\sin$ as in Eq.~(\ref{eq:3order_aff}), probably because the value of $\cos(\theta)$ grows monotonically with $\theta \in [0, \pi]$ but $\sin(\theta)$ does not.
%We set $\sigma_3=0.1$ in our implementation. 

\subsection{Further Discussions}
\subsubsection{Learning of problem structure}
The inherent working pattern of our proposed neural solver actually learns the underlying structure of graph matching problems. With the connection between graph matching problem and association graph, the original mathematical form of Lawler\textquotesingle s QAP transforms into a trackable structure with modern deep learning models. The problem structure is learned by GNN, resulting in a simplified Linear Assignment Problem solved with differentiable Sinkhorn algorithm. 
%\textcolor{red}{As will be shown in experiments, learning provides a cure when the objective function is severely biased in real-world scenarios.}
In \cite{KhalilNIPS17} some combinatorial problems over graphs are considered, where problems simplified by GNN are solved greedily. A similar scheme should generalize to other combinatorial problems, by exploiting the representative power of learning problem structures by deep learning.

\subsubsection{Matching-aware embedding} \label{sec:sk_emb}
The matching-aware Sinkhorn embedding is proposed to add one-to-one matching constraint at shallower embedding layers. Otherwise, the matching constraint is not considered until the output Sinkhorn layer. Early involvement of matching information has been proven effective for both learning-free (RRWM~\cite{ChoECCV10} vs SM~\cite{LeordeanuICCV05}) and learning based (PCA-GM vs PIA-GM~\cite{WangICCV19}) methods. We show the importance of Sinkhorn embedding by notable improvement in synthetic tests and ablation study on real-world images, additionally further improvement by introducing multi-head Sinkhorn embedding at the cost of increased computation.

\begin{table*}[tb!]
    \centering%Capability denotes whether this model is designed for two-graph matching i.e. QAP, hyper-graph matching (HGM) or multi-graph matching (MGM).
    \caption{Notation of all our proposed methods and their variants. }
    %\vspace{-10pt}
    %\resizebox{0.5\textwidth}{!}
    {
    \begin{tabular}{r|c|l}
    \hline
        model & capability & description \\
    \hline
        NGM/-v2 & classic QAP & Neural Graph Matching model introduced in Sec.~\ref{sec:ngm} /with enhanced feature in Sec.~\ref{sec:ngmv2}. \\
        %NGM+ & QAP & enhanced NGN with edge embedding in Sec.~\ref{sec:ngm+}. \\
        NHGM/-v2 & hypergraph matching & Neural Hyper-Graph Matching model introduced in Sec.~\ref{sec:nhgm} /with enhanced feature in Sec.~\ref{sec:ngmv2}. \\
        NMGM/-v2 & multi-graph matching & Neural Multi-Graph Matching model introduced in Sec.~\ref{sec:nmgm} /with enhanced feature in Sec.~\ref{sec:ngmv2}. \\
    \hline
        NGM-V & classic QAP & NGM by replacing Sinkhorn embedding in Eq.~(\ref{eq:sk_emb}) with vanilla embedding in Eq.~(\ref{eq:node}). \\
        NGM-MH &classic QAP & NGM model with multi-head Sinkhorn embedding (8 multi-head channels). \\
        NGM-SF & multi-graph matching & learned NGM model followed with learning-free spectral fusion. \\
        {NGM-G\texttt{X}} & classic QAP & sample \texttt{X} times by Gumbel-Sinkhorn and select the solution with the best objective. \\
        %NGM-SK & single permutation computed by Sinkhorn algorithm when inference (equivalent to NGM) \\
    \hline
    \end{tabular}
    }
    \label{tab:methods_notations}
\end{table*}

\subsubsection{Gumbel sampling for optimization problems}
\label{sec:gumbel_sk}
As a common post-processing step, the gap between doubly stochastic matrix $\mathbf{S}$ and permutation matrix $\mathbf{X}$ is fulfilled by Hungarian algorithm~\cite{HungarianMethod09} in a deterministic manner. From the probabilistic point of view, $\mathbf{S}$ represents a distribution on the space of permutation matrices, and our cross-entropy loss minimizes the distance between probability $\mathbf{S}$ and ground truth distribution $\mathbf{X}^{gt}$. Permutation with the highest probability is selected by Hungarian algorithm. 

Such a greedy scheme is empirically successful for matching. However, there might exist better solutions from the distribution, especially when optimizing the objective score of combinatorial problems. Therefore, we switch to Gumbel-Sinkhorn~\cite{menaICLR18learning} by replacing Eq.~(\ref{eq:sk_regularization}) with
\begin{equation}
    \label{eq:gumbel}
    \mathbf{s}_{ia}^{(k)} = \exp \left(\alpha_g (f_c(\mathbf{v}_{ia}^{(k)}) + g)\right)
\end{equation}
followed by Sinkhorn algorithm. Note the added $g$ is sampled from standard Gumbel distribution with cumulative distribution function (CDF):
\begin{equation}
    G(x)=e^{-e^{-x}}
\end{equation}
which models the distribution of extreme values from another distribution. By Eq.~(\ref{eq:gumbel}) sparser doubly-stochastic matrices can be sampled from the original distribution and repeated sampling provides a batch of samples. These sparse matrices are followed by Hungarian discretization, and the objective scores are computed and the best-performing solution is chosen as the final solution. Exploration and speed can be balanced by the number of Gumbel samples.

\section{Experiments}\label{sec:experiment}
%Extensive experiments have been conveyed with our proposed NGM and NHGM, compared to other state-of-the-art methods on solving QAP and learning affinity features for graph matching. 
%As our methods can work with either affinity matrix/tensor or raw RGB images as input, 
Experiments are conducted on a Linux workstation with Nvidia RTX8000 (48GB) GPU and Intel Xeon W-3175X CPU @ 3.10GHz with 128GB RAM. 

We test Lawler\textquotesingle s QAP in two settings: i) synthetic point registration, which takes affinity matrix/tensor as input, and ii) QAPLIB with large-scale real-world QAP instances where the network learns to minimize the objective score. We also test our method for a vision application of Lawler\textquotesingle s QAP in the sense that CNN features of image keypoints are learned and matched on real images. For keypoint matching, matching accuracy is computed as the percentage of correct matchings among all true matchings.
%We further specify our methods by suffix ``EM'' and ``SKEM'' 
%NGM-EM (NHGM-EM for hypergraph matching) and NGM-SKEM (NHGM-SKEM) 
%for the version whereby sole embedding and Skinhorn embedding is performed, respectively. See discussions between Eq.~(\ref{eq:sk_emb}) and Eq.~(\ref{eq:classification}). 

We also perform hypergraph and multiple graph matching tests to evaluate our NHGM/-v2 and NMGM/-v2. Our PyTorch implementation of NGM/-v2, NHGM/-v2, NMGM/-v2 involves a three-layer GNN, with graph feature channels $l_1=l_2=l_3=16$. Other hyperparameters are set as $\alpha =\hat{\alpha}= 20, \lambda_2 = 1, \lambda_3=1.5$. We set batch size=8. NGM, NHGM, NMGM are trained with SGD and 0.9 Nesterov momentum~\cite{ICML13nesterov}, and the v2 models are trained with Adam optimizer~\cite{adam}. Detailed learning rate configurations can be found in the following part of this section. Tab.~\ref{tab:methods_notations} summarizes all our methods and their variants.

%Experimental result reveals the superiority of our methods in matching performance, as well as its robustness to deformation and outliers.

\subsection{Synthetic Experiment for QAP Learning}
\subsubsection{Protocol setting}
In the synthetic experiment, sets of random points in the 2D plane are matched by comparison with other competitive learning-free graph matching solvers. For each trial, we generate 10 sets of ground truth points whose coordinates are in the plane $U(0,1) \times U(0,1)$. Synthetic points are distorted by random scaling from $U(1-\delta_s, 1+\delta_s)$ and additive random noise $N(0, \sigma_n^2)$. From each ground truth set, 200 graphs are sampled for training and 100 for testing, resulting in totally 2,000 training samples and 1,000 testing samples in one trial. We assume graph structure is unknown to the GM solver, therefore we construct the reference graph by Delaunay triangulation, and the target graph (may contain outliers) is fully connected. Outliers are also randomly sampled from $U(0,1) \times U(0,1)$. By default there are 10 inliers without outlier, with $\delta_s=0.1, \sigma_n=0$. We construct the same affinity matrix to formulate Lawler\textquotesingle s QAP for all methods. 

%All results are obtained on the same synthetic dataset for fair comparison. 

\begin{figure*}[tb!]
  \centering
  \includegraphics[width=0.97\textwidth]{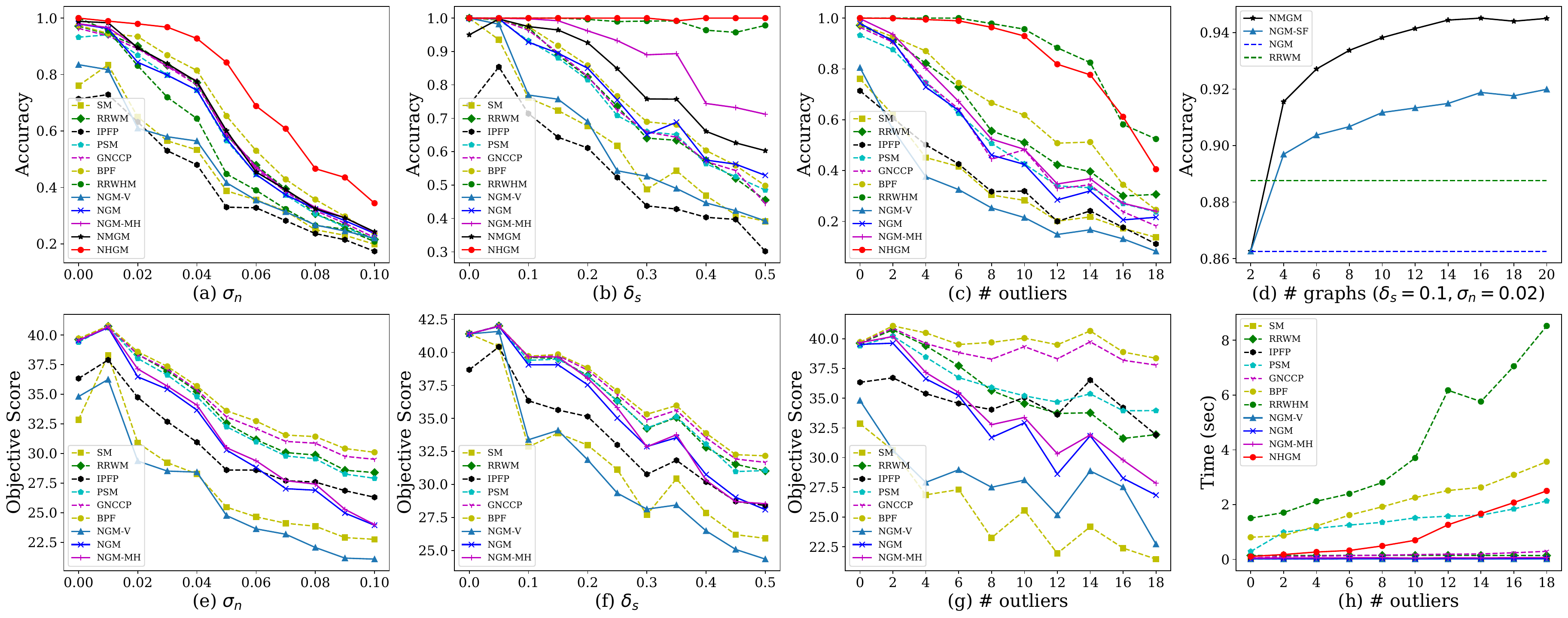}
  %\vspace{-20pt}
  \caption{Synthetic test by varying deformation level $\sigma_n, \delta_s$, number of outliers/graphs. Note it learns a QAP solver based on the given affinity matrix/tensor, without learning any affinity model. This feature is not supported in GMN~\cite{ZanfirCVPR18} and PCA-GM \cite{WangICCV19} thus they cannot be compared.}
  %\vspace{-10pt}
  \label{fig:synthetic}
\end{figure*}

\subsubsection{Peer methods}
As existing learning methods~\cite{ZanfirCVPR18,WangICCV19,ZhangICCV19} cannot handle learning with Lawler\textquotesingle s QAP with a given affinity matrix, here we first compare learning-free methods: \textbf{1)~SM}~\cite{LeordeanuICCV05} considers graph matching as discovering graph cluster by spectral numerical technique; \textbf{2)~RRWM}~\cite{ChoECCV10} adopts a random-walk view with reweighted jump on graph matching; \textbf{3)~IPFP}~\cite{LeordeanuNIPS09} iteratively improves a given solution via integer projection; \textbf{4)~PSM}~\cite{EgoziPAMI13} improves SM via a probabilistic view; \textbf{5)~GNCCP}~\cite{LiuPAMI12} is a convex-concave path-following method for graph matching and \textbf{6)~BPF}~\cite{WangPAMI17} improves path following techniques by branch switching, reaching state-of-the-art performance on graph matching. Additionally, hyper-graph matching algorithm \textbf{7)~RRWHM}~\cite{LeeCVPR11} extending powerful RRWM to hyper-graph scenarios is also compared. Second-order affinity is integrated into third-order tensor for RRWHM following \cite{LeeCVPR11}. In this experiment, second-order affinity is modeled by $K_{ia,jb}=\exp\left({-(\mathbf{f}_{ij}-\mathbf{f}_{ab})^2} / {\sigma^2_2}\right)$ where $\mathbf f_{ij}$ is edge length $E_{ij}$. We empirically set $\sigma_2^2=5\times 10^{-7}$ for all experiments. The third-order affinity model follows Eq.~(\ref{eq:3order_aff}) with $\sigma_3=0.1$. 

For fair comparison of run time, we re-implement the parallelization-friendly SM, RRWM and RRWHM solvers with GPU which can be more scalable than their original single-thread version on CPU. While the other compared methods involve iterative computing and complicated branching, which are not suitable for GPU. Thus the CPU version  released by \cite{WangPAMI17} are compared. NGM-V means vanilla NGM without Sinkhorn embedding (see discussion between Eq.~(\ref{eq:node}) and Eq.~(\ref{eq:sk_emb})) and NGM-MH means multi-head Sinkhorn embedding with NGM, by concatenating additional 8 Sinkhorn channels to $\mathbf{m}^{(k)}$ in Eq.~(\ref{eq:sk_emb}). The smoothing technique \cite{PachauriNIPS13} is adopted for multi-matching baseline NGM-SF, where learned NGM is followed with spectral fusion.
Since NGM/NHGM/NMGM-v2 share the same solver module with NGM/NHGN/NMGM, they are not compared in synthetic test. The learning rate starts at $10^{-2}$ decays by 10 every 5,000 steps. Multi-graph matching involves 4 graphs by default. The Hungarian algorithm is used as the common discretization step.% in graph matching. 

\subsubsection{Result and discussions}
Fig.~\ref{fig:synthetic}(a-c) shows our proposed NGM performs comparatively with state-of-the-art solvers in matching accuracy, and can even surpass under severe random scaling Fig.~\ref{fig:synthetic}(b). Further improvement in accuracy is achieved via multi-head Sinkhorn embedding model NGM-MH. NMGM gains steadily from NGM by fusing multi-matching information, and in Fig.~\ref{fig:synthetic}(d) we show the improvement in NMGM by introducing more graphs, and the necessity of learning joint matching as NMGM steadily outperforms NGM-SF whose weights are from two-graph NGM. With the third-order affinity, NHGM shows state-of-the-art robustness to noise, scaling and outliers. Compared to learning-free hyper-graph matching RRWHM~\cite{LeeCVPR11}, our NHGM performs comparatively in the precense of outliers as shown in Fig.~\ref{fig:synthetic}(c). While it performs more robustly to noises and scaling in Fig.~\ref{fig:synthetic}(a\&b). As shown in Fig.~\ref{fig:synthetic}(h), NGM, NGM-V and NGM-MH are among the fastest graph matching algorithms, and due to efficient GPU parallelization, NHGM is comparatively fast against PSM~\cite{EgoziPAMI13} and more time-efficient compared to state-of-the-art BPF~\cite{WangPAMI17}. In contrast, traditional hypergraph matching algorithms e.g.\ RRWHM~\cite{LeeCVPR11} are usually much slower than second-order graph matching and unscalable to larger size of problems.

We report the QAP objective score solved by two-graph matching methods in Fig.~\ref{fig:synthetic}(e-g), where interestingly our learning-based solvers reach relatively low scores compared to their corresponding accuracy. As have been discussed in Sec.~\ref{sec:intro}, the QAP objective may be biased under noisy conditions i.e. the optimal solution to QAP may not correspond to true matching. Our solvers learns to ignore the noisy patterns in input affinity matrix. Such a phenomenon becomes more severe in matching real-world images, and the strength of learning-based solvers becomes more significant, as will be shown in Sec.~\ref{sec:exp_voc} in details.

The effectiveness of matching-aware Sinkhorn embedding is shown in the accuracy gap between NGM and NGM-V. Further improvement is achieved by multi-head Sinkhorn embedding in NGM-MH, especially with random scaling in Fig.~\ref{fig:synthetic}(b). As discussed in Sec.~\ref{sec:sk_emb}, NGM-V without Sinkhorn embedding works in a way similar to SM as the embedding procedure does not consider the assignment constraint, and they also perform closely to each other. On the other hand, by exploiting Sinkhorn embedding, NGM and NGM-MH are conceptually similar to RRWM as all of them try to incorporate the assignment constraint on the fly. Their performances are also very close. 

%Fig.~\ref{fig:synthetic} shows our proposed NGM surpasses SM and reaches comparative results against the powerful learning-free method RRWM. The robustness of hyper matching NHGM becomes significant when we further disturb the matching problem by random scaling (Fig.~\ref{fig:synthetic}(b)) and outliers (Fig.~\ref{fig:synthetic}(c)). In Fig.~\ref{fig:synthetic}(a)(b), NMGM obtains steady improvement by fusing multi-graph information, and in Fig.~\ref{fig:synthetic}(d) we show the improvement in NMGM by introducing more graphs, and the necessity of learning joint matching as NMGM steadily outperforms NMGM-T whose weights are directly transferred from two-graph NGM. In fact, NGM-V without Sinkhorn embedding works in a way similar to SM as the embedding procedure does not consider the assignment constraint, and they also perform closely to each other. On the other hand, by exploiting Sinkhorn embedding, NGM is conceptually similar to RRWM as both methods try to incorporate the assignment constraint on the fly. Interestingly their performances are also very close. 

%Such technique falls into a simple but effective formulation which is solved by computing the leading eigenvector of affinity matrix.

\begin{figure*}[tb!]
    \centering
    \subfigure[Normalized objective score (lower is better) of our best-performing NGM-G5k against learning-free QAP solvers. Failed instances are plotted at the top of y-axis. The instances are firstly split into two parts based on whether NGM-G5K performs better than Sinkhorn-JA~\cite{KushinskySIAM19} or not, and then sorted by the normalized score of NGM-G5k. NGM-G5k surpasses state-of-the-art learning-free method Sinkhorn-JA~\cite{KushinskySIAM19} on 85 out of 133 instances.]{
        \includegraphics[width=\textwidth]{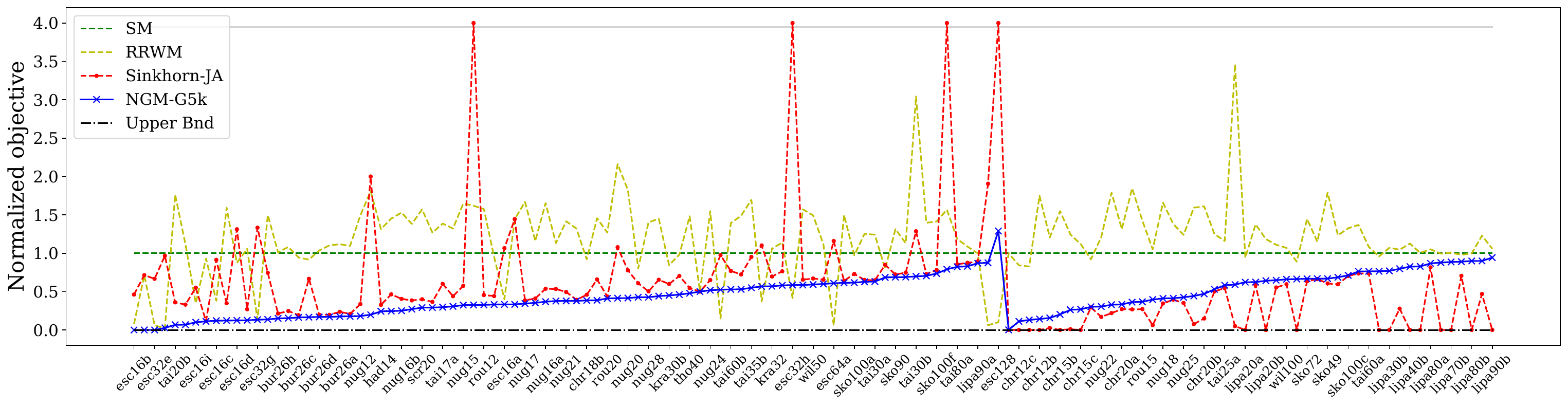}
        \label{fig:qaplib_main}
    }
    \subfigure[Normalized objective score (lower is better) comparing different sampling settings. More samples in Gumbel-Sinkhorn guarantee a higher probability of finding better solutions, at the cost of increased computation. NGM always picks the permutation matrix with the highest probability with Sinkhorn and Hungarian algorithms, and performs similarly to NGM-G50 and NGM-G500. The order is kept in line with Fig.~\ref{fig:qaplib_main}.]{
        \includegraphics[width=\textwidth]{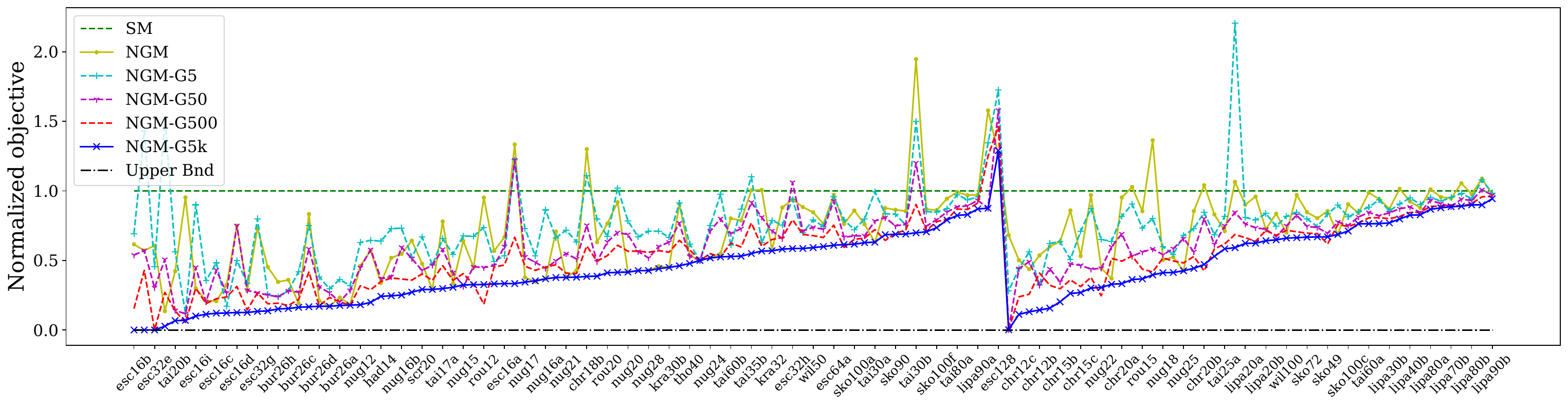}
        \label{fig:qaplib_ngms}
    }
    %\vspace{-10pt}
    \caption{Normalized objective score on real-world QAPLIB instances. Only half of the instance labels are shown on x-axis due to limited space.}        

    %\vspace{-10pt}
\end{figure*}

\begin{figure}[tb!]
    \centering
    \includegraphics[width=0.36\textwidth]{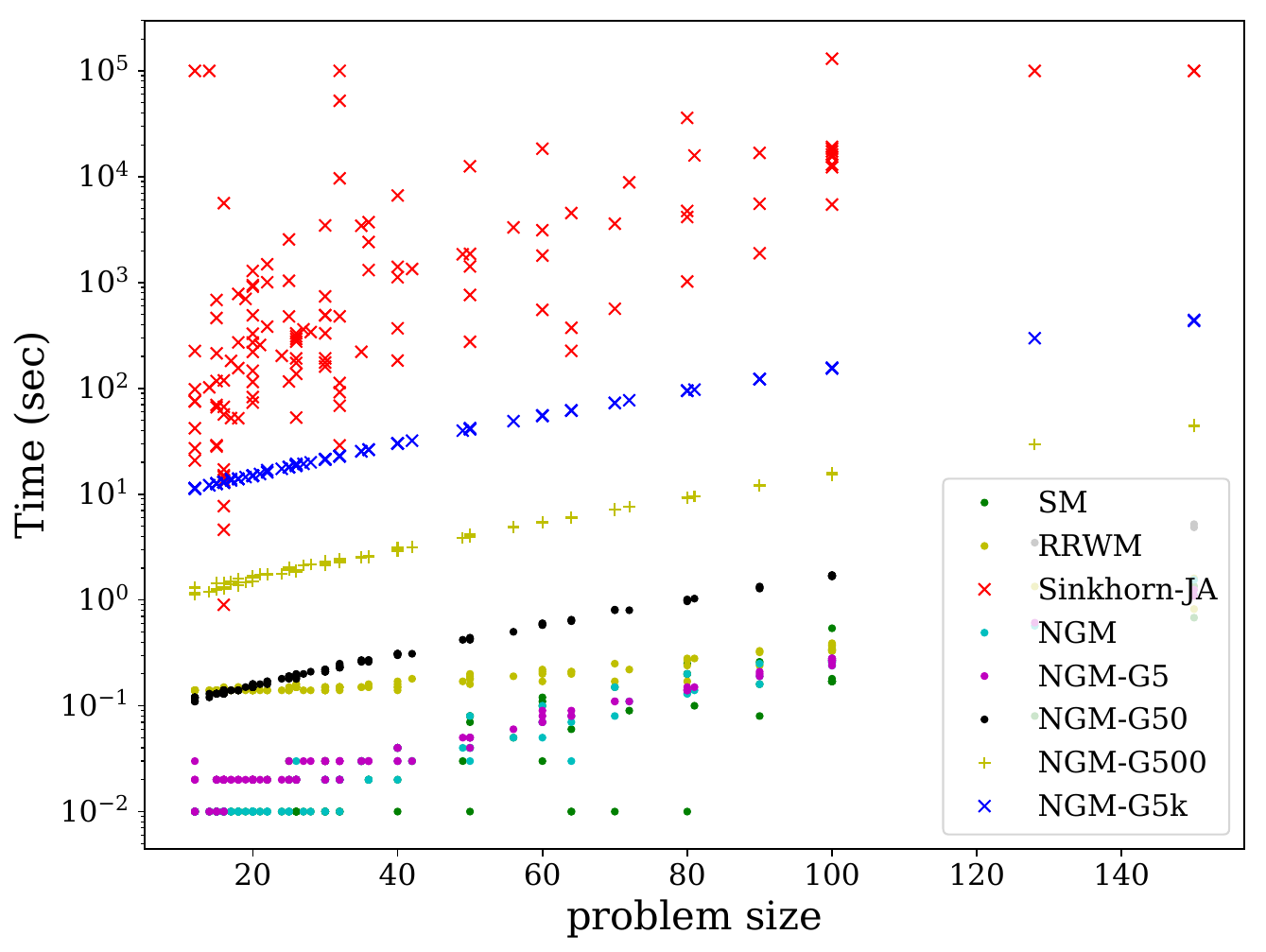}
     %\vspace{-10pt}
    \caption{Run time (log-scale) against problem size i.e. number of nodes for each graph. All methods except Sinkhorn-JA are implemented and executed on GPUs, as Sinkhorn-JA is challenging to be parallelized.}
    \label{fig:qaplib_time}
\end{figure}

\subsection{Learning Real-world QAP Instances} \label{sec:exp_qaplib}
%QAP is a fundamental formulation of many real-world problems including graph matching.
\subsubsection{Experiment setting}
Our NGM solves the most general Lawler\textquotesingle s QAP, which has a wide range of applications beyond vision. Evaluation on QAPLIB~\cite{Burkard1997QAPLIB} is performed to show the capability of NGM on learning the QAP objective score, which should be minimized in QAPLIB (in contrast, objective score is maximized in graph matching). The QAPLIB contains 134 real-world QAP instances from 15 categories, e.g.\ planning a hospital facility layout~\cite{hahn2001hospital}. The problem size is defined as $n_1=n_2=n$ from Lawler\textquotesingle s QAP in Eq.~(\ref{eq:lawler_qap}), and ranges from 12 to 256. Results are reported on 133 instances with $12 \leq n \leq 150$, as the most challenging \texttt{tai256c} is computationally intractable with our testbed (275GB GPU memory is required for intermediate computing). We set the loss function as the objective score of QAP, keeping the model architecture unchanged. We formulate a optimization task where the objective score is minimized:
\begin{equation}
    L_{obj} = \mathrm{vec}(\mathbf{S})^{\top} \mathbf{K} \mathrm{vec}(\mathbf{S})
\end{equation}
where $\mathbf{S}$ is from the output Sinkhorn layer of NGM. Given optimal $L_{obj}$ is reached, the learned $\mathbf{S}$ is a double-stochastically relaxed solution to original QAP.
To explore the feasible space, Gumbel-Sinkhorn discussed in Sec.~\ref{sec:gumbel_sk} is adopted during inference.

The naming of QAPLIB instances are based on the following rule: the prefix is the problem category which is the name of the author who proposes the problem, and the number means the size of the problem. If there are multiple problems with the same size, QAPLIB appends a letter starting from \texttt{a} to distinguish.
\begin{equation}
    \qaplibbur_{\text{author name}} \text{---} \qaplibtwentysix_{\text{problem size}} \text{---} \qapliba_{\text{index (optional)}} \notag
\end{equation}
We train one network for each set of problems with the same prefix (i.e.\ problems from the same category), because they usually have common structures (e.g.\ all \texttt{bur} problems are keyboard layout design problems). And the problem sizes may vary for problems in the same category (e.g.\ the sizes of \texttt{esc} problems vary from 16 to 128). 

We train one model for each category, and report the normalized objective score. In consideration of compact and intuitive illustration, the normalized objective score is computed with the upper bound (primal bound) provided by the up-to-date online benchmark\footnote{\href{http://anjos.mgi.polymtl.ca/qaplib/inst.html}{http://anjos.mgi.polymtl.ca/qaplib/inst.html}} and normalized by the baseline solver spectral matching (SM)~\cite{LeordeanuICCV05}:
\begin{equation}
    norm\_score = \frac{solved\_score - upper\_bnd}{SM\_score - upper\_bnd}
\end{equation}
Detailed per-instance scores and timing statics are available in Appendix. Both standard NGM model (NGM) and NGM with different Gumbel sampling numbers (NGM-G\texttt{X}, \texttt{X} = number of samples) are validated for their performance. The learning rate is initialized at $10^{-4}$ and decays by 10 every 50,000 steps. Batch size is set to 1 and the regularization of Gumbel Sinkhorn $\alpha_g=1$. Our proposed methods are compared fairly with our GPU implementation of RRWM~\cite{ChoECCV10} and SM~\cite{LeordeanuICCV05}, and results provided in the paper of Sinkhorn-JA~\cite{KushinskySIAM19} (runs on Intel Xeon CPU @ 2.40 GHz). For the problem instances not reported in \cite{KushinskySIAM19}, we assume Sinkhorn-JA fails to reach any feasible solution, as there is no explanation of missing instances in the original paper.

\begin{table*}[tb!]
    \caption{Best-performing occurrence count on QAPLIB among all instances and all tested solvers. Our NGM-G5k surpasses all competing methods on most categories and best performs on 72 out of 133 instances.}
     %\vspace{-10pt}
    \label{tab:qaplib_bestnum}
    \centering
    % Table generated by Excel2LaTeX from sheet 'Sheet5'
\begin{tabular}{r|ccccccccccccccc|c}
\hline
     category& {bur} & {chr} & {els} & {esc} & {had} & {kra} & {lipa} & {nug} & {rou} & {scr} & {sko} & {ste} & {tai} & {tho} & {wil} & {total}  \\
      \hline
\#instances & 8     & 14    & 1     & 19    & 5     & 3     & 16    & 15    & 3     & 3     & 13    & 3     & 25    & 3     & 2     & 133 \\
\hline
SM~\cite{LeordeanuICCV05}    & 0     & 0     & 0     & 1     & 0     & 0     & 0     & 0     & 0     & 0     & 0     & 0     & 0     & 0     & 0     & 1 \\
RRWM~\cite{ChoECCV10}  & 0     & 0     & 0     & 9     & 0     & 0     & 0     & 0     & 0     & 0     & 0     & 0     & 0     & 0     & 0     & 9 \\
Sinkhorn-JA~\cite{KushinskySIAM19} & 0     & \textbf{14} & \textbf{1} & 1     & 0     & 0     & \textbf{15} & 4     & \textbf{1} & 0     & 1     & 0     & 7     & 1     & \textbf{1} & 46 \\
NGM & 0     & 0     & 0     & 1     & 0     & 2     & 0     & 5     & 0     & 0     & 0     & \textbf{3} & 1     & 0     & 0     & 12 \\
NGM-G5 & 0     & 0     & 0     & 1     & 0     & 0     & 0     & 0     & 0     & 0     & 0     & 0     & 0     & 0     & 0     & 1 \\
NGM-G50 & 0     & 0     & 0     & 1     & 0     & 0     & 0     & 1     & 0     & 0     & 0     & 0     & 0     & 0     & 0     & 2 \\
NGM-G500 & 1     & 0     & 0     & 3     & 0     & 0     & 0     & 1     & \textbf{1} & 0     & 5     & 0     & 1     & 0     & 0     & 12 \\
NGM-G5k & \textbf{8} & 1     & 0     & \textbf{11} & \textbf{5} & \textbf{3} & 1     & \textbf{10} & \textbf{1} & \textbf{3} & \textbf{7} & 2     & \textbf{17} & \textbf{2} & \textbf{1} & \textbf{72} \\
\hline
\end{tabular}%
\end{table*}

\subsubsection{Result and analysis}
In Fig.~\ref{fig:qaplib_main}, our method beats RRWM~\cite{ChoECCV10} and SM~\cite{LeordeanuICCV05} and is comparative and even superior against state-of-the-art Sinkhorn-JA~\cite{KushinskySIAM19}. As there is no learning-based QAP solver, only non-learning methods are compared. The effectiveness of Gumbel sampling discussed in Sec.~\ref{sec:gumbel_sk} is validated in Fig.~\ref{fig:qaplib_ngms}, where Gumbel-based NGM-G5k consistently outperforms deterministic NGM, which always picks the permutation with the highest probability by Hungarian algorithm. With decreased sampling number, the performance of Gumbel-based methods gradually degenerates. It suggests that more exploration over sampling space guarantees higher expectations on better solutions. 
%We also show in Fig.~\ref{fig:qaplib_cm} that the learned QAP solver generalizes soundly to unseen problem instances, where models are trained by problems on column and tested on row. The problems are arbitrarily selected. Objective scores are plotted and darker color denotes better performance.
%QAPLIB statistics with 
\begin{table*}[tb!]
    \centering
    \caption{Pearson correlation coefficient $r$ (only for $|r| \geq 0.2$ are shown) between the listed statistics for each problem instance in QAPLIB and the corresponding $gap$ of two methods (see Eq.~(\ref{eq:gap})), as a signal of problem difficulty (bigger gap more difficult). Higher value denotes increased negative effect of the corresponding statistics to the final solution quality. The correlation between statistics and the difference of two methods is also listed on the last row, where higher value denotes NGM-G5k is more sensitive than Sinkhorn-JA and vice versa.}% Note that relatively small $r$ ($|r| < 0.2$) is usually considered no relation between two statistics.
     %\vspace{-10pt}
    % Table generated by Excel2LaTeX from sheet 'Sheet1'
    \renewcommand\arraystretch{1.3}
    \begin{tabular}{r|cccccccc}
    \hline
    method  & $nz$ & $nz/n^4$ (sparsity) & $\mathbf{K}_{std}/\mathbf{K}_{max}$ & $d_{min}/\bar{\mathbf{K}}$ & $d_{max}/\bar{\mathbf{K}}$ & $d_{std}/\bar{\mathbf{K}}$ & $d_{max}/\bar{d}$ & $d_{std}/\bar{d}$ \\
    \hline
    NGM-G5k & 0.130  & \textbf{0.631} & \textbf{0.212} & \underline{\textbf{-0.222}} & \textbf{0.230} & \textbf{0.286} & \textbf{0.230} & \textbf{0.280} \\
    Sinkhorn-JA & \textbf{0.270} & \textbf{0.545} & -0.090  & \underline{\textbf{-0.220}} & \textbf{0.291} & \textbf{0.355} & \textbf{0.291} & \textbf{0.475} \\
    \hline
    $gap_{\text{NGM}} - gap_{\text{SJA}}$ & -0.184  & -0.121  & \textbf{0.234} & 0.071  & -0.137  & -0.163  & -0.137  & \underline{\textbf{-0.288}} \\
    \hline
    \end{tabular}%
    \label{tab:qaplib_correlation}
\end{table*}

\begin{figure}[tb!]
    \centering
    \includegraphics[width=0.28\textwidth]{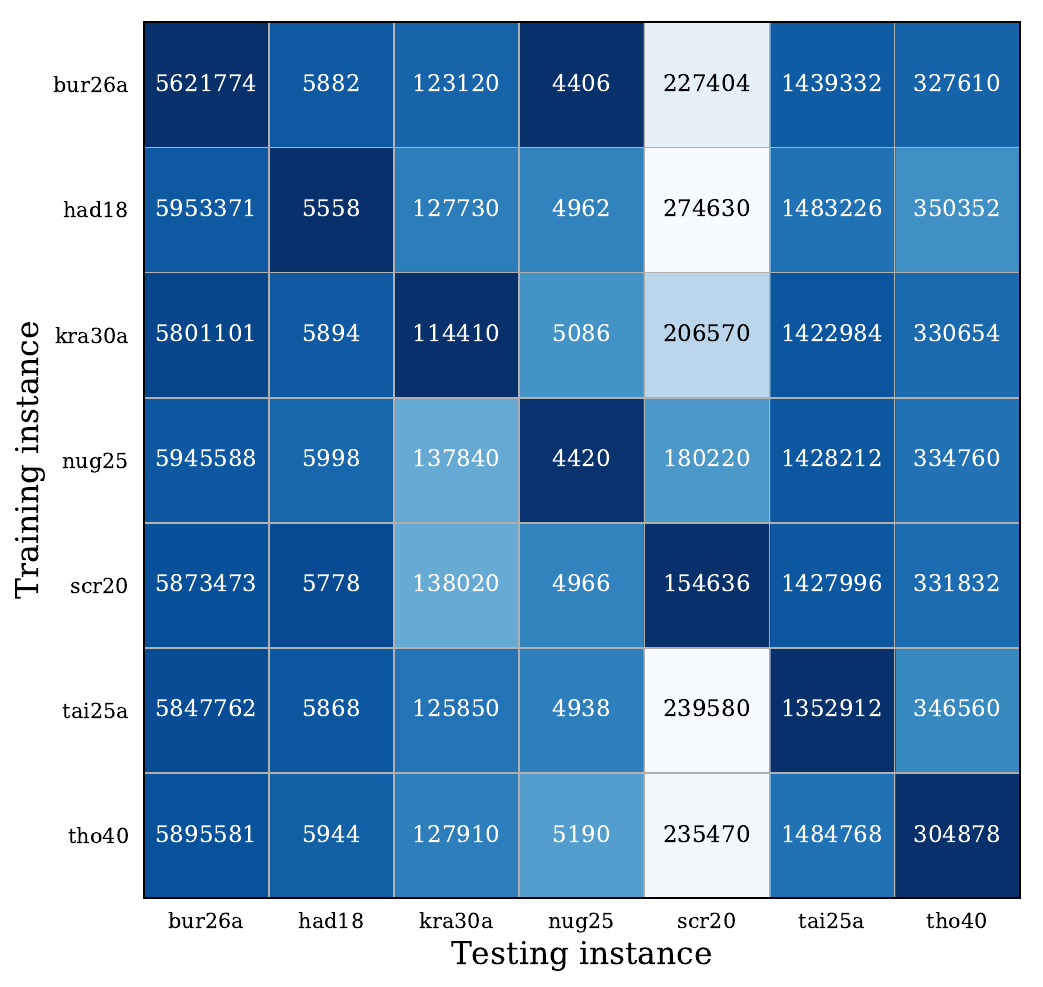}
    %\vspace{-10pt}
    \caption{Generalization test by confusion matrix cross QAP problem instances, where models are learned with instances on y-axis and tested with instances on x-axis. Darker color and lower score correspond to better performance. The tasks are randomly selected from the QAPLIB benchmark. Problem sizes are shown by the numbers in instance names, and our method is insensitive to problem sizes. }
    \label{fig:qaplib_cm}
\end{figure}

Further evaluation is given in Tab.~\ref{tab:qaplib_bestnum} and Fig.~\ref{fig:qaplib_time}. With learning and Gumbel sampling, our  NGM-G5k finds the best solution among 72 out of 133 instances, while state-of-the-art learning-free solver Sinkhorn-JA~\cite{KushinskySIAM19} outperforms on 46 instances so that learning-based solvers e.g.\ NGM can fit a wider range of problems compared to traditional solvers. More importantly, our best-performing model NGM-G5k is of a magnitude faster than Sinkhorn-JA, and adjusting the sampling number of Gumbel method enables balancing between solution quality and computational demand. Finally, we show the generalization ability among different instances by confusion matrix in Fig.~\ref{fig:qaplib_cm}, where NGM-G5k is trained and tested on different randomly picked instances. Our model generalizes soundly to unseen instances with different problem sizes. In conclusion, our learning QAP solvers achieve the best accuracy-speed trade-off on QAPLIB and can generalize among different problems.

\subsubsection{Further discussion}
As shown in Tab.~\ref{tab:qaplib_bestnum} and Fig.~\ref{fig:qaplib_main}, learning-free Sinkhorn-JA~\cite{KushinskySIAM19} and our NGM-G5k performs better on separate categories of QAPLIB, e.g.\ Sinkhorn-JA performs better on \texttt{chr} and \texttt{lipa} instances while our method is more powerful on \texttt{bur}, \texttt{esc}, \texttt{nug}, \texttt{scr} and \texttt{tai}. Some statistical studies are conducted to discover the relation between model behavior and problem patterns, shedding light for future research on both learning-based and learning-free solvers.

For each problem instance, some statistics are summarized from each instance\textquotesingle s affinity matrix: problem size $n$, mean value $\bar{\mathbf{K}}$, minimum value $\mathbf{K}_{min}$, maximum value $\mathbf{K}_{max}$, standard deviation $\mathbf{K}_{std}$, number of zeros $nz$, mean degree (of association graph) $\bar{d}$, minimum degree $d_{min}$, maximum degree $d_{max}$ and standard deviation of degree $d_{std}$. The performance of algorithms is represented by the ${gap}$ of the solved objective score against upper bound:
\begin{equation}\label{eq:gap}
    gap=\frac{solved\_score - upper\_bnd}{solved\_score}
\end{equation}
It means the percentage of improvement can be made compared to best-known optima (usually solved at extremely high complexity). Pearson correlation coefficients $r$ are computed between each ${gap}$ and corresponding statistics, additionally some meaningful combinations of the statistics. Items with $|r|\geq0.2$ are listed in Tab.~\ref{tab:qaplib_correlation}, where positive correlation means a negative effect on solver\textquotesingle s performance because a lower $gap$ is better. The correlation between problem statistics and the difference between two methods are also reported.

\begin{figure*}[tb!]
    \centering
    \includegraphics[width=\textwidth]{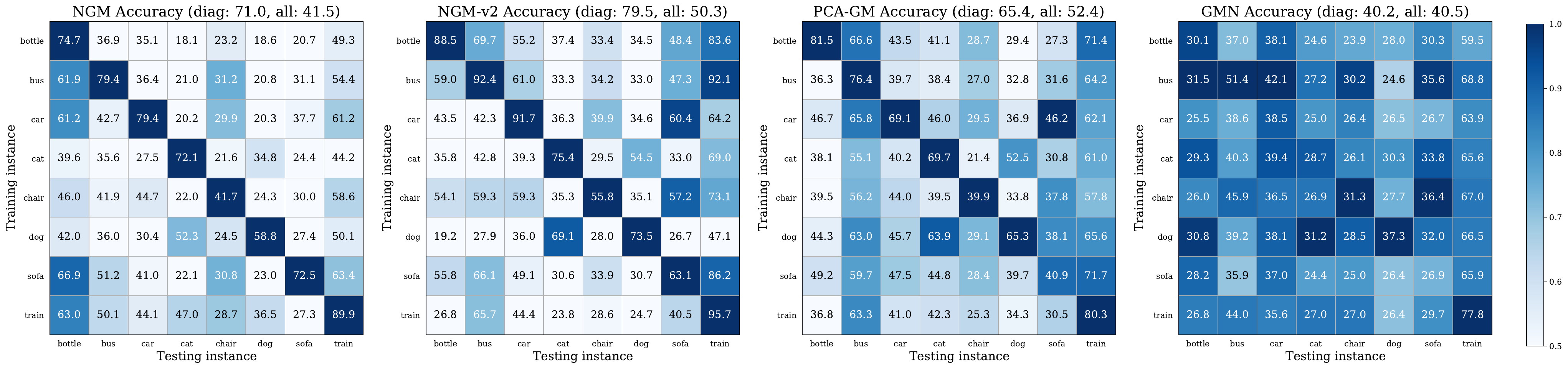}
    %\vspace{-25pt}
    \caption{Generalization study shown as confusion matrix of our proposed NGM and NGM-v2 in line with \cite{WangICCV19,ZanfirCVPR18} on Pascal VOC Keypoint. Models are trained on categories on the y-axis and tested on categories on the x-axis. Darker color denotes relatively better performance in the same column, and the averaged accuracy on both diagonal part (learned during training) and all elements (trained+generalized) of confusion matrices are reported in the brackets over confusion matrices. The train/test split follows the main experiment and eight categories are selected randomly.}
    %\vspace{-10pt}
    \label{fig:cm_voc}
\end{figure*}

\begin{table*}[tb!]
    \centering
    %\vspace{8pt}
    \caption{Matching accuracy (\%) on Pascal VOC Keypoint. Best results are in bold.}
     %\vspace{-10pt}
    \resizebox{\textwidth}{!}
    {
        \setlength{\tabcolsep}{2pt}
        
        % Table generated by Excel2LaTeX from sheet 'PascalVOC'
        \begin{tabular}{r|cccccccccccccccccccc|c}
        \hline
        method & aero  & bike  & bird  & boat  & bottle & bus   & car   & cat   & chair & cow   & table & dog   & horse & mbkie & person & plant & sheep & sofa  & train & tv    & mean \\\hline
        GMN~\cite{ZanfirCVPR18} & 41.6  & 59.6  & 60.3  & 48.0  & 79.2  & 70.2  & 67.4  & 64.9  & 39.2  & 61.3  & 66.9  & 59.8  & 61.1  & 59.8  & 37.2  & 78.2  & 68.0  & 49.9  & 84.2  & 91.4  & 62.4 \\
        PCA-GM~\cite{WangICCV19} & 49.8  & 61.9  & 65.3  & 57.2  & 78.8  & 75.6  & 64.7  & 69.7  & 41.6  & 63.4  & 50.7  & 67.1  & 66.7  & 61.6  & 44.5  & 81.2  & 67.8  & 59.2  & 78.5  & 90.4  & 64.8 \\
        IPCA-GM~\cite{WangPAMI20} & 53.8  & 66.2  & 67.1  & 61.2  & 80.4  & 75.3  & 72.6  & 72.5  & 44.6  & 65.2  & 54.3  & 67.2  & 67.9  & 64.2  & 47.9  & 84.4  & 70.8  & 64.0  & 83.8  & 90.8  & 67.7 \\
        CIE-H~\cite{YuICLR20} & 49.9  & 63.1  & 70.7  & 53.0  & 82.4  & 75.4  & 67.7  & 72.3  & 42.4  & 66.9  & 69.9  & 69.5  & 70.7  & 62.0  & 46.7  & 85.0  & 70.0  & 61.8  & 80.2  & 91.8  & 67.6 \\
        LCS~\cite{WangCVPR20} & 46.9  & 58.0  & 63.6  & 69.9  & \textbf{87.8} & 79.8  & 71.8  & 60.3  & 44.8  & 64.3  & 79.4  & 57.5  & 64.4  & 57.6  & 52.4  & 96.1  & 62.9  & 65.8  & 94.4  & 92.0  & 68.5 \\
        BBGM~\cite{RolinekECCV20} & \textbf{61.9} & 71.1  & \textbf{79.7} & 79.0  & 87.4  & 94.0  & \textbf{89.5} & 80.2  & 56.8  & 79.1  & 64.6  & \textbf{78.9} & 76.2  & 75.1  & \textbf{65.2} & 98.2  & 77.3  & \textbf{77.0} & 94.9  & \textbf{93.9} & 79.0 \\
        \hline
        NGM~(ours) & 50.1  & 63.5  & 57.9  & 53.4  & 79.8  & 77.1  & 73.6  & 68.2  & 41.1  & 66.4  & 40.8  & 60.3  & 61.9  & 63.5  & 45.6  & 77.1  & 69.3  & 65.5  & 79.2  & 88.2  & 64.1 \\
        NHGM~(ours) & 52.4  & 62.2  & 58.3  & 55.7  & 78.7  & 77.7  & 74.4  & 70.7  & 42.0  & 64.6  & 53.8  & 61.0  & 61.9  & 60.8  & 46.8  & 79.1  & 66.8  & 55.1  & 80.9  & 88.7  & 64.6 \\
        NGM-v2~(ours) & 61.8  & 71.2  & 77.6  & 78.8  & 87.3  & 93.6  & 87.7  & 79.8  & 55.4  & 77.8  & \textbf{89.5} & 78.8  & \textbf{80.1} & \textbf{79.2} & 62.6  & 97.7  & 77.7  & 75.7  & 96.7  & 93.2  & 80.1 \\
        NHGM-v2~(ours) & 59.9  & \textbf{71.5} & 77.2  & \textbf{79.0} & 87.7  & \textbf{94.6} & 89.0  & \textbf{81.8} & \textbf{60.0} & \textbf{81.3} & 87.0  & 78.1  & 76.5  & 77.5  & 64.4  & \textbf{98.7} & \textbf{77.8} & 75.4  & \textbf{97.9} & 92.8  & \textbf{80.4} \\\hline
        \end{tabular}%
    }
    \label{tab:voc_main}
\end{table*}

In the first two columns in Tab.~\ref{tab:qaplib_correlation}, the higher sparsity ($nz/n^4$, proportion of zeros in affinity matrix) makes the problem significantly more challenging for both NGM-G5k and Sinkhorn-JA, and the same conclusion holds for a larger number of zeros $nz$. Then the normalized standard deviation $\mathbf{K}_{std}/\mathbf{K}_{max}$ shows some weak negative effect for NGM-G5k, but little correlation to Sinkhorn-JA. In the last five columns both methods are affected by higher degrees in association graph, while Sinkhorn-JA seems more sensitive to the normalized standard deviation of degrees $d_{std}/\bar{\mathbf{K}}, d_{std}/\bar{d}$. In summary, sparsity is the key challenge for both NGM-G5k (where message passing paths are blocked) and Sinkhorn-JA (where it becomes harder to find tight lower bounds). Furthermore, learning with the association graph can restrain the noisy deviation in degrees. Future improvement may be achieved by designing graph learning models with higher capacity, and designing global communication mechanisms against sparse association graphs.

\subsection{Real Image for Joint CNN and QAP Learning}
%\subsubsection{Protocol Setting}
Our matching net allows for raw image input, from which a CNN is learned (see Fig.~\ref{fig:network_structure}). We evaluate semantic keypoint matching on Pascal VOC dataset with Berkeley annotations\footnote{https://www2.eecs.berkeley.edu/Research/Projects/CS/vision/ shape/poselets/voc2011\_keypoints\_Feb2012.tgz}~\cite{bourdev2009poselets_VOCkeypoint} and Willow ObjectClass dataset~\cite{ChoICCV13}. 

\subsubsection{Results on Pascal VOC Keypoint dataset} \label{sec:exp_voc}
This natural image dataset~\cite{bourdev2009poselets_VOCkeypoint} consists of 20 instance classes with keypoint labels. We follow~\cite{WangICCV19}, where image pairs with inlier positions are fed into the model. We consider it a challenging dataset because instance may vary from its scale, pose and illumination, and the number of keypoints in each image varies from $6$ to $23$. The shallow learning method HARG-SSVM~\cite{ChoICCV13} incorporates a fix-sized reference graph, therefore it is inapplicable to our experiment setting where instances from the same category have different inliers. As our multi-matching mechanism in Sec.~\ref{sec:nmgm} requires the same number of nodes among graphs, our multi-graph model NMGM and NMGM-v2 are not compared either.

In line with the protocol of ~\cite{ZanfirCVPR18,WangICCV19}, we filter out those poorly annotated images which are meaningless for matching. Then we get 7,020 training samples and 1,682 testing samples. Instances are cropped around their ground truth bounding boxes and resized to $256 \times 256$ before fed into the network. As discussed in Sec.~\ref{sec:model_CNN4imgs}, we adopt VGG16 backbone~\cite{simonyanICLR14vgg} and construct the affinity matrix from the same CNN layers: \texttt{relu4\_2} for node features and \texttt{relu5\_1} for edge features. 
%Edge features are built by concatenating feature vectors of starting node and ending node. The affinity score is modeled by weighted inner product of two vectors, containing learnable weights. 
%Readers are referred to \cite{ZanfirCVPR18} for the comprehensive graph construction pipeline.
The learning rate starts at $10^{-2}$ and decays by $10$ every 10,000 steps. For the two input images, one graph is constructed by Delaunay triangulation and the other is fully-connected. $\sigma_3=10^{-4}$ is set for third-order affinity of NHGM.

\begin{figure*}[tb!]
    \centering
    \includegraphics[width=\textwidth]{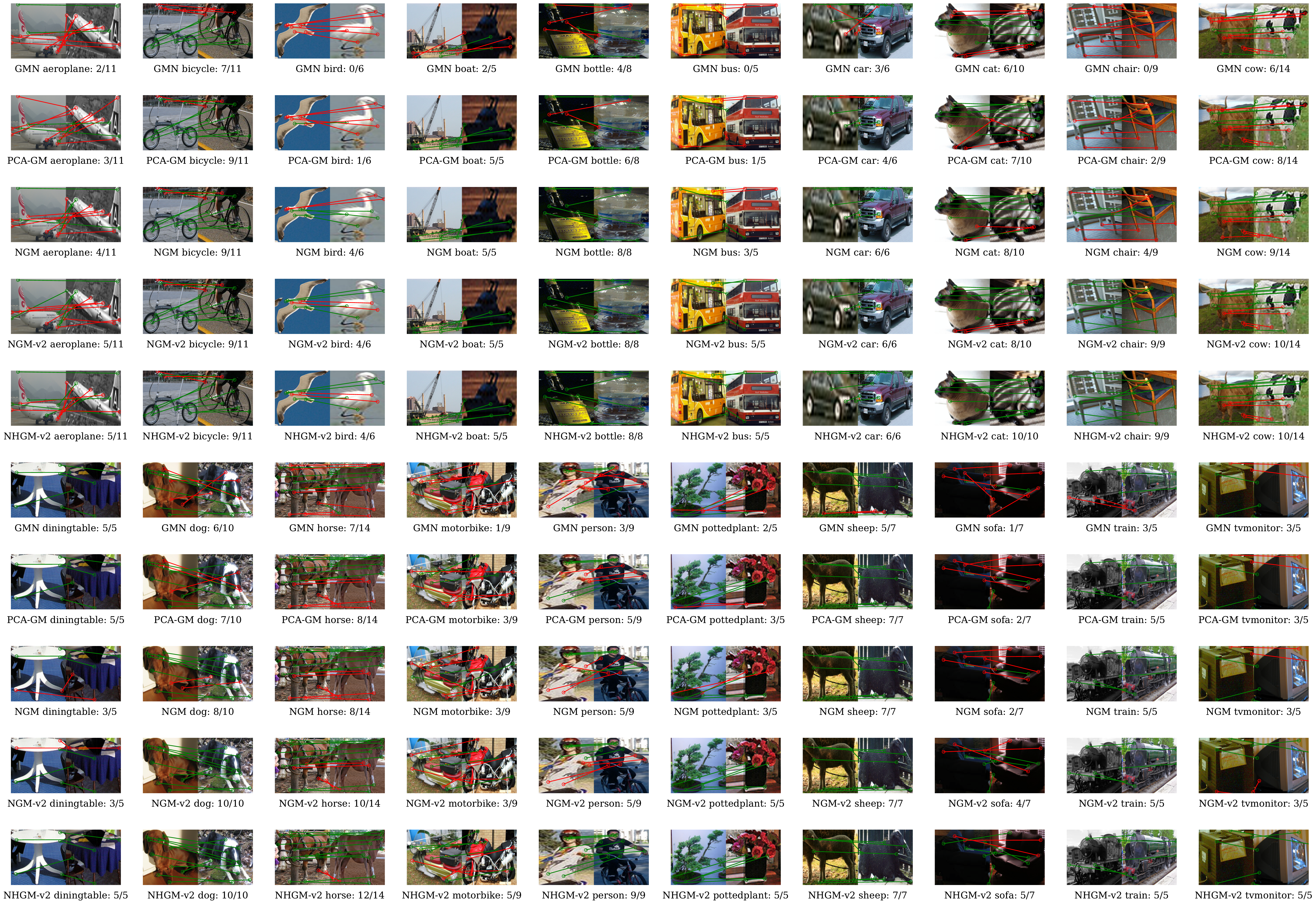}
    %\vspace{-20pt}
    \caption{Visualization of matching results on 20 Pascal VOC Keypoint categories. Green and red represent correct and incorrect predictions, respectively. As many instances vary a lot in their pose and appearance, this dataset is considered challenging even for deep graph matching.}
    \label{fig:visualization}
    %\vspace{-10pt}
\end{figure*}

For NGM/NHGM-v2, the feature extractor is replaced by the enhanced backbone with SplineConv and weighted inner-product affinity as discussed in Sec.~\ref{sec:ngmv2}, while keeping other modules unchanged. $\sigma_3=0.1$ is set for third-order affinity of NHGM-v2. Following \cite{RolinekECCV20}, we set learning rate $2\times10^{-3}$ for VGG16 and $2\times10^{-5}$ for other modules.

We compare \textbf{GMN}~\cite{ZanfirCVPR18}, \textbf{PCA-GM}~\cite{WangICCV19}, \textbf{LCS}~\cite{WangCVPR20}, \textbf{BBGM}~\cite{RolinekECCV20}, by which affinity functions are learned for graph matching. It is worth noting that we discover the experiment setting implemented by BBGM~\cite{RolinekECCV20} is easier than ours, because they filter out keypoints which are out of the bounding box but we do not. Therefore, we reimplement BBGM~\cite{RolinekECCV20} to fit our experiment setting and the reported BBGM results in this paper are slightly worse than the results in its original paper. 

Results in Tab.~\ref{tab:voc_main} show that with CNN feature and QAP solver learned jointly, our methods surpass competing methods on most categories, especially best performs in terms of mean accuracy. Specifically, NGM surpasses deep graph matching method PCA-GM, and it is worth noting that PCA-GM incorporates explicit modeling on higher-order and cross-graph affinities, while only second-order affinity is considered in our QAP formulation (and third-order for hypergraph matching). With enhanced feature extractor, our NGM-v2 surpasses state-of-the-art BBGM in terms of accuracy, which is not surprising because NGM-v2 and BBGM share the same feature extractor, and NGM-v2 learns both feature extractor and the graph matching solver, but BBGM only learns the feature extractor. NHGM and NHGM-v bring further improvement with respect to NGM and NGM-v2 by exploiting hypergraph affinities. 
The GPU memory cost and running speed during training are listed: NGM (4761MB, 13.7 pairs/s); NGM-v2 (5707MB, 14.1 pairs/s); NHGM (9251MB, 10.1 pairs/s); NHGM-v2 (38060MB, 11.0 pairs/s). NGM and NGM-v2 are comparable to PCA-GM~\cite{WangICCV19} (5165MB, 14.4 pairs/s).

%Our proposed NGM, NHGM, NGM+ are compared with RRWM~\cite{ChoECCV10}, GMN~\cite{ZanfirCVPR18}, PCA-GM~\cite{WangICCV19}.

%ImgNet-RRWM builds affinity matrix from pretrained weights from ImageNet and solves QAP with learning-free RRWM solver~\cite{ChoECCV10}, ImgNet+NGM adopts ImageNet CNN weights but takes the solver trained with NGM, and NGM+RRWM's affinity matrix is learned by NGM and QAP solved via RRWM~\cite{ChoECCV10}. The performance of ImageNet-VGG degenerates significantly proving the importance of learning distinct CNN features for graph matching. RRWM outperforms the spectral-matching~\cite{LeordeanuICCV05} based learning scheme GMN, but fails to surpass our joint CNN and QAP learning NGM. 

The generalization ability is further validated by confusion matrices for NGM and NGM-v2 as shown in Fig.~\ref{fig:cm_voc}. Models are trained only with categories on the x-axis and tested with all categories. The color map is determined by the accuracy in current cell normalized by the highest accuracy in its column. As shown in the confusion matrices, both NGM and NGM-v2 own some generalization ability between visually similar categories, e.g.\ chair and sofa, cat and dog, bus and train. Compared to peer deep graph matching methods \cite{WangICCV19,ZanfirCVPR18}, NGM and NGM-v2 fit better on the training categories on the diagonal of confusion matrices, while on the other hand NGM and NGM-v2 seem less powerful than PCA-GM when considering all categories, since most categories are irrelevant to the training category.

The matching visualization is given in Fig.~\ref{fig:visualization}. The error patterns of NGM-v2, NHGM-v2 are similar but differ from NGM and PCA-GM, because they are based on different feature extractors. As can be seen from the difference between NGM and NGM-v2, better feature extractor can promise better matching accuracy in various categories. And NHGM-v2 improves NGM-v2 by correcting some existing errors, e.g. in ``cat'', ``motorbike'', ``person'' and ``tvmonitor''.

%In general, our methods are more powerful on rigid categories, e.g.\ aeroplane, bike, car, motorbike, pottedplant and sofa. The node-wise and edge-wise affinity formulation is helpful in modeling structural information, as shown in visualization of ``car''. In contrast, PCA-GM seems more powerful on non-rigid objects such as birds and horses. Our methods, however, fail when the objects vary a lot in their pose and appearance (see ``bird'' and ``bus''). Additionally, as there exist many non-rigid and pose-varying objects, the effect of adding geometric-based higher-order information in NHGM seems not significant. In contrast, NGM+ provides more convincing result, probably due to the increased model capacity with edge embedding.

\begin{table*}[t]
    \centering
    \caption{Controlled experiment by replacing NGM with unlearned ImageNet CNN and RRWM solver. The last row denotes our joint CNN-solver learning method NGM which outperforms.}
    %\vspace{-10pt}
    \resizebox{\textwidth}{!}{
    \setlength{\tabcolsep}{2pt}
    \begin{tabular}{c|c|cccccccccccccccccccc|c}
        \hline
        CNN & solver & aero  & bike  & bird  & boat  & bottle & bus   & car   & cat   & chair & cow   & table & dog   & horse & mbkie & person & plant & sheep & sofa  & train & tv    & mean \\\hline
        ImgNet VGG16~\cite{simonyanICLR14vgg} & RRWM~\cite{ChoECCV10} & 16.1  & 22.3  & 20.8  & 21.8  & 21.3  & 31.0  & 23.2  & 25.4  & 18.6  & 20.5  & 20.6  & 21.7  & 18.8  & 21.9  & 13.5  & 28.6  & 21.7  & 18.3  & 50.5  & 42.8  & 24.0 \\
        ImgNet VGG16~\cite{simonyanICLR14vgg} & NGM solver~(ours) & 30.8  & 42.5  & 44.3  & 33.8  & 39.8  & 52.2  & 49.2  & 53.9  & 27.5  & 42.4  & 29.3  & 49.1  & 45.1  & 45.1  & 24.0  & 48.3  & 49.9  & 29.9  & 70.2  & 73.3  & 44.0 \\
        NGM VGG16~(ours) & RRWM~\cite{ChoECCV10} & 41.5  & 54.7  & 54.3  & 50.3  & 67.9  & 74.3  & 70.3  & 60.6  & 42.3  & 59.1  & 48.1  & 57.3  & 59.1  & 56.2  & 40.6  & 69.6  & 63.1  & 52.2  & 76.3  & 87.8  & 59.3 \\
        \hline
        NGM VGG16~(ours) & NGM solver~(ours) & 50.1  & 63.5  & 57.9  & 53.4  & 79.8  & 77.1  & 73.6  & 68.2  & 41.1  & 66.4  & 40.8  & 60.3  & 61.9  & 63.5  & 45.6  & 77.1  & 69.3  & 65.5  & 79.2  & 88.2  & 64.1 \\
        \hline
    \end{tabular}
    }
         %\vspace{-10pt}

    \label{tab:voc_controlled}
\end{table*}

\begin{table}[tb!]
\centering
\caption{Matching accuracy (\%) on Willow ObjectClass dataset. The first 5 rows denote learning-based peer methods. Row 6-9 show our two-graph and hypergraph matching methods. Row 10-11 indicate learning-free multi-graph matching baselines, and HiPPI~\cite{BernardICCV19} jointly matches 40, 50, 109, 40, 66 graphs respectively for 5 Willow categories. The last 2 rows denote our two multi-graph matching learning variants.}
 %\vspace{-10pt}
\resizebox{0.48\textwidth}{!}
    {
% Table generated by Excel2LaTeX from sheet 'Willow'
\begin{tabular}{r|c|ccccc|c}
\hline
model & \# graphs & car   & duck  & face  & m-bike & w-bottle & mean \\\hline
GMN~\cite{ZanfirCVPR18} & 2     & 67.9  & 76.7  & 99.8  & 69.2  & 83.1  & 79.3 \\
PCA-GM~\cite{WangICCV19} & 2     & 87.6  & 83.6  & \textbf{100.0} & 77.6  & 88.4  & 87.4 \\
IPCA-GM~\cite{WangPAMI20} & 2     & 90.4  & 88.6  & \textbf{100.0} & 83.0  & 88.3  & 90.1 \\
BBGM~\cite{RolinekECCV20} & 2     & 96.8  & 89.9  & \textbf{100.0} & 99.8  & \textbf{99.4} & 97.2 \\
LCS~\cite{WangCVPR20} & 2     & 91.2  & 86.2  & \textbf{100.0} & 99.4  & 97.9  & 94.9 \\
\hline
NGM~(ours) & 2     & 84.2  & 77.6  & 99.4  & 76.8  & 88.3  & 85.3 \\
NHGM~(ours) & 2     & 86.5  & 72.2  & 99.9  & 79.3  & 89.4  & 85.5 \\
NGM-v2~(ours) & 2     & 97.4  & 93.4  & \textbf{100.0} & 98.6  & 98.3  & 97.5 \\
NHGM-v2~(ours) & 2     & 97.4  & 93.9  & \textbf{100.0} & 98.6  & 98.9  & 97.8 \\
\hline
HiPPI~\cite{BernardICCV19} & $\geq$40   & 74.0  & 88.0  & \textbf{100.0 } & 84.0  & 95.0  & 88.2 \\
MGM-Floyd~\cite{JiangPAMI20} & 32    & 85.0  & 79.3  & \textbf{100.0 } & 84.3  & 93.1  & 88.3 \\
%NGM-SF~(ours + \cite{PachauriNIPS13}) & 10    & 90.9  & 79.4  & 99.6  & 83.1  & 85.2  & 87.6 \\
\hline
NMGM~(ours) & 10    & 78.5  & 92.1  & \textbf{100.0} & 78.7  & 94.8  & 88.8 \\
NMGM-v2~(ours) & 10    & \textbf{97.6} & \textbf{94.5} & \textbf{100.0} & \textbf{100.0} & 99.0  & \textbf{98.2} \\
\hline
\end{tabular}%
}
%\vspace{-12pt}
\label{tab:willow_main}
 %\vspace{-10pt}
\end{table}

\subsubsection{Results on Willow ObjectClass dataset} 
This natural image dataset covers 5 categories. Each category contains at least 40 images, and all instances in the same class share 10 distinctive image keypoints. We mainly evaluate multi-graph matching learning of NMGM and NMGM-v2 on Willow ObjectClass. Following the protocol in \cite{WangICCV19}, we directly train our methods on the first 20 images and report testing results on the rest. The learning rate starts at $10^{-2}$ and decays by 10 every 500 steps.

%The suffix in ``NMGM-$k$'' means learning joint matching among $k$ graphs, and suffix ``T'' means transferring NGM weights by adding a permutation synchronization head without learning on joint matching. 
The performance of HARG-SSVM~\cite{ChoICCV13}, GMN~\cite{ZanfirCVPR18} and PCA-GM~\cite{WangICCV19} reported in \cite{WangICCV19} are listed and compared. 
%Learning-free spectral fusion multi-matching algorithm~\cite{PachauriNIPS13} is also integrated and tested as NGM-SF, by post-processing on NGM pairwise matching.
Two novel multi-graph matching algorithms HiPPI~\cite{BernardICCV19} and MGM-Floyd~\cite{JiangPAMI20} are considered as learning-free baselines.
%, and it's worth noting that GMN and PCA-GM incorporate additional training data from Pascal VOC Keypoint. Pretraining on Pascal VOC Keypoint is not adopted because we empirically find our model overfit on VOC data, and 
Tab.~\ref{tab:willow_main} shows that NGM and NHGM performs comparatively to PCA-GM especially on rigid objects, and NGM-v2 and NHGM-v2 surpass the state-of-the-art BBGM~\cite{RolinekECCV20}. Our NMGM involves less number of graphs than learning-free methods HiPPI~\cite{BernardICCV19} and MGM-Floyd~\cite{JiangPAMI20}, but achieves higher accuracy thanks to end-to-end multi-graph matching learning. Our NMGM-v2 best performs among all methods.

%and NHGM performs comparatively to PCA-GM especially on rigid objects, and NMGM surpasses PCA-GM by learning multi-graph information. Learning joint matching on more graphs can achieve further improvement, as NMGM with 6 graphs outperforms 3 graphs. However, the multi-matching capacity seems saturated as there is little improvement by involving 16 graphs. With no learning on multi-graph matching, little improvement is observed for NGM-SF from 3 graphs to 6 graphs and 16 graphs, except the accuracy on motorbikes. The WILLOW dataset is relatively small for deep graph matching as there are only $5\times20=100$ training images. 

\subsubsection{Ablation study on Pascal VOC Keypoint dataset}
We validate the effectiveness of learning on both CNN and the solver by a controlled experiment. In Tab.~\ref{tab:voc_controlled}, the first entry denotes whether the CNN weights are obtained from pretrained ImageNet classifier (ImgNet)~\cite{deng2009imagenet} or learned graph matching model (NGM); the second entry means either learning-free solver RRWM or learning-based solver NGM is adopted to solve QAP. The necessity of learning on both CNN and the solver is validated, and our joint CNN and solver learning method best performs among them. Learning with CNN unsurprisingly mitigates the gap between classification task and matching task, while it is worth noting the learned solver is nearly twice accurate against RRWM with ImageNet CNN (44.0\% vs 24.0\%), showing the robustness on our solver side against noises.

%As synthetic test results have already shown the effectiveness of Sinkhorn embedding, further 
Ablation study is performed on our proposed modules in NHGM-v2 on Pascal VOC Keypoint dataset, as shown in Tab.~\ref{tab:ablation}. The baseline model is built following NGM introduced in Sec.~\ref{sec:ngm}, but node affinity is ignored in model input, i.e. $\mathbf{v}_{ia}^{(0)}=1$ and Sinkhorn embedding (Sec.~\ref{sec:sk_emb}) is excluded. The effectiveness of node affinity, Sinkhorn embedding, and NGM-v2\textquotesingle s SplineConv feature and weighted-inner product affinity are validated by adding these components successively. Finally, we add hypergraph affinity proposed in Sec.~\ref{sec:v2_hypergraph} for NHGM-v2. Tab.~\ref{tab:ablation} shows that SplineConv feature contributes significantly to the matching accuracy, and NGM can extend seamlessly to NGM-v2 because our method handles the most general form of QAP.

%Based on the NGM model with node affinity and Sinkhorn embedding (64.1\%), we also test other novel edge-embedding schemes including EGNN(C)~\cite{gongCVPR19exploiting} and HyperConv~\cite{JiangIJCAI19}, and it is discovered not all edge-embedding models are suitable for learning with the complicated association graph.

%\subsection{Experiments on Multiple Graph Matching}
\begin{table}[tb!]
    \centering
    \caption{Ablation study of NHGM-v2 on Pascal VOC Keypoint dataset.}
    %\vspace{-10pt}
    %\resizebox{0.48\textwidth}{!}
    {
        \begin{tabular}{r|rr}
            \hline
            model & accuracy & relative acc \\
            \hline
            baseline NGM & 58.7\% & \\
            + node affinity & 59.6\% & +0.9\% \\
            + Sinkhorn embedding  & 64.1\% & +4.5\% \\
            + SplineConv feature & 74.3\% & +10.2\%\\
            + weighted inner-product affinity & 80.1\% & +5.8\%\\
            + hypergraph affinity & 80.4\% & +0.3\% \\
            %\hline[0.2pt]
            \hline%[0.2pt]
            %edge embedding & EGNN(C)~\cite{gongCVPR19exploiting} & HyperConv~\cite{JiangIJCAI19} & Ours Eq.~(\ref{eq:ngm+_edge})\\
            %\hline
            %accuracy & 53.3\% & 55.9\% & \textbf{66.1\%} \\
            %\hline
        \end{tabular}
    }
   % \vspace{-10pt}
    \label{tab:ablation}
    %\vspace{-12pt}
     %\vspace{-10pt}
\end{table}
\section{Conclusion and Outlook}\label{sec:conclusion}
We have presented a novel neural graph matching network, with three main highlights: i) The first graph matching network directly learning Lawler’s QAP which is general with a wide range of applications e.g.\ on QAPLIB beyond visual matching. This is in contrast to many existing works that can only take separate graphs as input. ii) The first deep network for hypergraph matching which involves third-order edges. iii) The first network for deep learning of multiple graph matching. Extensive experimental results on synthetic and real-world data show the state-of-the-art performance of our approach. In particular, it shows the notable cost-efficiency advantages against learning-free methods.

In future work, we will explore more scalable approaches for handling large-scale QAP problems. For graph matching, it indicates more graphs with more nodes for matching. For its generality, we will also apply our model to more application areas beyond computer vision.

\ifCLASSOPTIONcompsoc
  % The Computer Society usually uses the plural form
  \section*{Acknowledgments}
\else
   %regular IEEE prefers the singular form
  \section*{Acknowledgment}
\fi
The work is partially supported by National Key Research and Development Program of China (2020AAA0107600), Shanghai Municipal Science and Technology Major Project (2021SHZDZX0102), NSFC (61972250, U19B2035).
%The authors are thankful to the anonymous reviewers and editor's valuable suggestions to improve the paper. 

% Can use something like this to put references on a page
% by themselves when using endfloat and the captionsoff option.
\ifCLASSOPTIONcaptionsoff
  \newpage
\fi

% trigger a \newpage just before the given reference
% number - used to balance the columns on the last page
% adjust value as needed - may need to be readjusted if
% the document is modified later
%\IEEEtriggeratref{8}
% The "triggered" command can be changed if desired:
%\IEEEtriggercmd{\enlargethispage{-5in}}

% references section

% can use a bibliography generated by BibTeX as a .bbl file
% BibTeX documentation can be easily obtained at:
% http://mirror.ctan.org/biblio/bibtex/contrib/doc/
% The IEEEtran BibTeX style support page is at:
% http://www.michaelshell.org/tex/ieeetran/bibtex/
%\bibliographystyle{IEEEtran}
% argument is your BibTeX string definitions and bibliography database(s)
%\bibliography{IEEEabrv,../bib/paper}
%
% <OR> manually copy in the resultant .bbl file
% set second argument of \begin to the number of references
% (used to reserve space for the reference number labels box)

\bibliographystyle{IEEEtran}
\bibliography{IEEEabrv,survey.bib}
\begin{IEEEbiography}[{\includegraphics[width=1in,height=1.25in,clip,keepaspectratio]{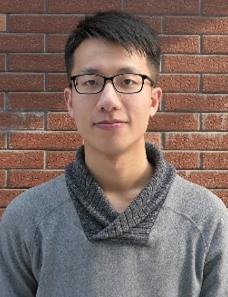}}]{Runzhong Wang} (S\textquotesingle 21) is currently a PhD Candidate with Department of Computer Science and Engineering, and AI Institute, Shanghai Jiao Tong University. He obtained B.E. in Electrical Engineering from Shanghai Jiao Tong University. He has published a series of first-authored papers in ICCV 2019, NeurIPS 2020, CVPR 2021, and IEEE TPAMI on machine learning for combinatorial optimization. He serves a reviewer for CVPR 2020/2021, NeurIPS 2020 and AAAI 2021. His research interests include machine learning and combinatorial optimization. He has open-sourced and is maintaining the deep graph matching solver available at: \href{https://github.com/Thinklab-SJTU/PCA-GM}{https://github.com/Thinklab-SJTU/PCA-GM} which has received more than 400 stars.
\end{IEEEbiography}
%\vspace{-20pt}
\begin{IEEEbiography}[{\includegraphics[width=1in,height=1.25in,clip,keepaspectratio]{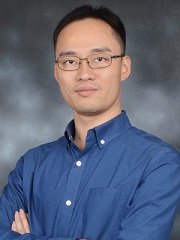}}]{Junchi Yan} (S\textquotesingle 10-M\textquotesingle 11-SM\textquotesingle 21) is currently an Associate Professor with Shanghai Jiao Tong University, Shanghai, China. Before that, he was a Senior Research Staff Member and Principal Scientist with IBM Research -- China, where he started his career in April 2011. He obtained the Ph.D. in Electrical Engineering from Shanghai Jiao Tong University. His research interests include machine learning and computer vision. He serves as Area Chair for ICPR 2020, CVPR 2021 and Senior PC for CIKM 2019, IJCAI 2021.
\end{IEEEbiography}
\begin{IEEEbiography}[{\includegraphics[width=1in,height=1.25in,clip,keepaspectratio]{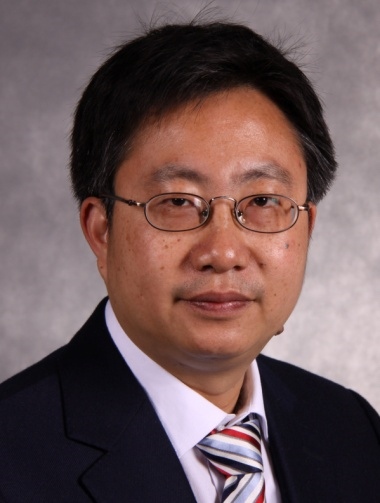}}]{Xiaokang Yang} (M\textquotesingle 00-SM\textquotesingle 04-F\textquotesingle 19) received the B. S. degree from Xiamen University, in 1994, the M. S. degree from Chinese Academy of Sciences in 1997, and the Ph.D. degree from Shanghai Jiao Tong University in 2000. He is currently a Distinguished Professor of School of Electronic Information and Electrical Engineering, Shanghai Jiao Tong University, Shanghai, China. His research interests include visual signal processing and communication, media analysis and retrieval, and pattern recognition. He serves as an Associate Editor of IEEE Transactions on Multimedia, IEEE Signal Processing Letters. He is a Fellow of IEEE.
\end{IEEEbiography}
\end{document}